\PassOptionsToPackage{table}{xcolor}
\documentclass{article}
\usepackage{iclr2027_conference,times}
\iclrfinalcopy
\usepackage[utf8]{inputenc}
\usepackage[T1]{fontenc}
\usepackage{enumitem}
\usepackage{url}
\usepackage{booktabs}
\usepackage{amsmath,amsfonts}
\usepackage{microtype}
\usepackage[table]{xcolor}
\usepackage{graphicx}
\usepackage{float}
\usepackage{placeins}
\usepackage{tabularx,array}
\usepackage{tikz}
\usepackage{fontawesome5}
\usepackage{tcolorbox}
\usepackage{hyperref}
\newcolumntype{Y}{>{\raggedright\arraybackslash}X}
\renewcommand{\arraystretch}{1.08}
\definecolor{catMathNT}{HTML}{C96F1F}
\definecolor{catDynProg}{HTML}{15756A}
\definecolor{catGraph}{HTML}{8E5BA6}
\definecolor{catGreedy}{HTML}{B48A3A}
\definecolor{catImpl}{HTML}{7C93C0}
\definecolor{catOther}{HTML}{5E9C8C}
\definecolor{catDataStr}{HTML}{2B5FAD}
\newcommand{\catMath}{\textcolor{catMathNT}{\faCalculator}}
\newcommand{\catDP}{\textcolor{catDynProg}{\faTable}}
\newcommand{\catGR}{\textcolor{catGraph}{\faProjectDiagram}}
\newcommand{\catGS}{\textcolor{catGreedy}{\faSearch}}
\newcommand{\catIS}{\textcolor{catImpl}{\faCogs}}
\newcommand{\catHash}{\textcolor{catOther}{\faHashtag}}
\newcommand{\catRange}{\textcolor{catDataStr}{\faBorderAll}}
\newcommand{\catSeq}{\textcolor{catMathNT}{\faIcon{exchange-alt}}}
\newcommand{\catSort}{\textcolor{catDynProg}{\faSortAmountDown}}
\newcommand{\catStr}{\textcolor{catGraph}{\faFont}}
\DeclareRobustCommand{\catlabel}[2]{%
  \ifmmode\text{#1\nobreak\hspace{0.22em}#2}%
  \else#1\nobreak\hspace{0.22em}#2\fi}
\definecolor{tabhead}{RGB}{223,232,244}
\definecolor{tabrule}{RGB}{120,140,165}
\newcommand{\tabsetupplain}{\setlength{\tabcolsep}{6pt}\arrayrulecolor{tabrule}}
\newcommand{\headrow}{\rowcolor{tabhead}}
\newcommand{\hd}[1]{\textbf{#1}}
\DeclareRobustCommand{\sixlegend}[2]{\begingroup\setlength{\fboxsep}{1pt}\colorbox[HTML]{#1}{#2}\endgroup}
\definecolor{findingbg}{HTML}{FFF8E7}
\definecolor{findingframe}{HTML}{DFC475}
\newtcolorbox{finding}{%
  colback=findingbg, colframe=findingframe,
  boxrule=0.55pt, arc=2pt,
  left=7pt, right=7pt, top=4pt, bottom=4pt,
  before skip=5pt, after skip=7pt}
\definecolor{pillfill}{RGB}{237,242,247}
\definecolor{pillline}{RGB}{176,196,214}
\newcommand{\pill}[1]{%
  \tikz[baseline=(pilltext.base)]{%
    \node[draw=pillline,fill=pillfill,rounded corners=3.5pt,%
          inner xsep=4pt,inner ysep=1.6pt,line width=0.3pt,%
          font=\scriptsize] (pilltext) {#1};}%
}
\newcommand{\pillsep}{\hspace{2pt}\allowbreak}

\title{\centering How code helps different tasks?\\A decompositional lens on LLM post-training}

\author{%
\begin{minipage}[t]{\dimexpr\textwidth-2\tabcolsep\relax}
\centering
\textbf{Zheng Yu}\textsuperscript{1,2,*}\quad
\textbf{Yiwei Li}\textsuperscript{3,*}\quad
\textbf{Yishen Chen}\textsuperscript{3}\quad
\textbf{Xiang Li}\textsuperscript{4}\\[3pt]
\textbf{Jiale Han}\textsuperscript{2}\quad
\textbf{Benyou Wang}\textsuperscript{3,\textdagger}\quad
\textbf{Jingbang Chen}\textsuperscript{3,2,\textdagger}\\[6pt]
{\normalfont\small
\textsuperscript{1}Sun Yat-sen University\\
\textsuperscript{2}Shenzhen Loop Area Institute\\
\textsuperscript{3}The Chinese University of Hong Kong, Shenzhen\\
\textsuperscript{4}Shenzhen Research Institute of Big Data\\[4pt]
\textsuperscript{*}Equal contribution.\quad
\textsuperscript{\textdagger}Corresponding authors.}
\end{minipage}%
}

\hypersetup{
  pdftitle={How code helps different tasks? A decompositional lens on LLM post-training},
  pdfauthor={Zheng Yu, Yiwei Li, Yishen Chen, Xiang Li, Jiale Han, Benyou Wang, Jingbang Chen},
  pdfsubject={Code data selection and cross-task transfer in LLM post-training}
}

\begin{document}
\maketitle
\lhead{Preprint}
\linespread{0.975}\selectfont
\begin{abstract}
Evaluating code data as a single corpus can obscure which types of code data benefit which models and downstream tasks. Effective data selection requires understanding both the benefits of individual categories and whether these benefits persist when categories are combined. We introduce a decompositional lens for studying these effects in LLM post-training. We first decompose an execution-verified code corpus into interpretable categories based on the computational patterns of its solutions. Through controlled fine-tuning experiments, we compare individual categories with a balanced mixture across instruction-tuned models on question answering, mathematics, and code generation. The resulting response maps reveal recurring gains in average question-answering performance, while the same category can improve one model or task and degrade another. The best-performing category also varies with the starting model and target task. We then compose compact mixtures guided by these results and examine whether benefits observed in individual categories persist under joint training. On selected model--task pairs, mixtures whose constituents each improve the target task outperform both their best constituent and full-corpus training while using roughly 10--15\% of the full corpus. These exploratory findings illustrate a \emph{less is more} pattern and highlight how the value of code data in post training depends on which categories are combined for which model and task.
\par\smallskip
\begingroup\hypersetup{hidelinks}\small
\noindent\makebox[\linewidth][c]{%
\href{https://github.com/William-Li123/code-generalization}{\faGithub\hspace{0.35em}\textbf{Code}}%
\hspace{1.5em}%
\href{https://huggingface.co/datasets/CodeLogic-Team/Code-generalization}{\raisebox{-0.15em}{\includegraphics[height=1.2em]{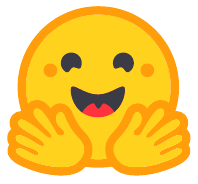}}\hspace{0.35em}\textbf{Dataset}}%
}
\endgroup
\end{abstract}

\section{Introduction}
Code data is an appealing source of supervision for language models because
its solutions express explicit computational procedures and can be checked
by execution. Previous studies have investigated its benefits beyond code
generation. \citet{aryabumi2024tocode} examine how code exposure and data
quality affect downstream performance during pre-training, while
\citet{zhang2025codingdata} study how the proportion of coding data in
instruction tuning affects reasoning across models and tasks. 
Recent work links these benefits to specific properties of the training
data. \citet{waheed2025codeinduced} find that structural perturbations can
affect downstream performance more strongly than semantic perturbations.
\citet{twist2026not} show that restricting training data to selected
structural-complexity ranges can outperform training on a broader range
of complexities. These findings motivate a finer-grained understanding of
which code examples provide useful supervision for a particular model
and downstream task.

This understanding has practical value for data selection.
Research on data mixture optimization and skill-based sampling shows that
selecting and combining training data can improve downstream performance
and training efficiency \citep{xie2023doremi,chen2023skillit}.
Targeted instruction tuning further demonstrates the value of selecting
examples that support specific downstream capabilities \citep{xia2024less}.
Within a code corpus, reference solutions express different computational
patterns, such as dynamic programming, hash-based counting, and string
parsing. Grouping examples by the computations performed in their solutions allows
us to compare the effects of distinct forms of code supervision across
models and tasks. Recombining these groups then tests whether the benefits
identified in individual categories extend to joint training.
Existing findings about code exposure, representation, and complexity do
not by themselves specify which computational categories benefit a given
model, or how their benefits change when categories are trained together.
Connecting category-specific effects to joint training could therefore
inform the construction of compact mixtures that improve performance
with less training data.

\begin{figure}[!t]
\centering
\includegraphics[width=\textwidth]{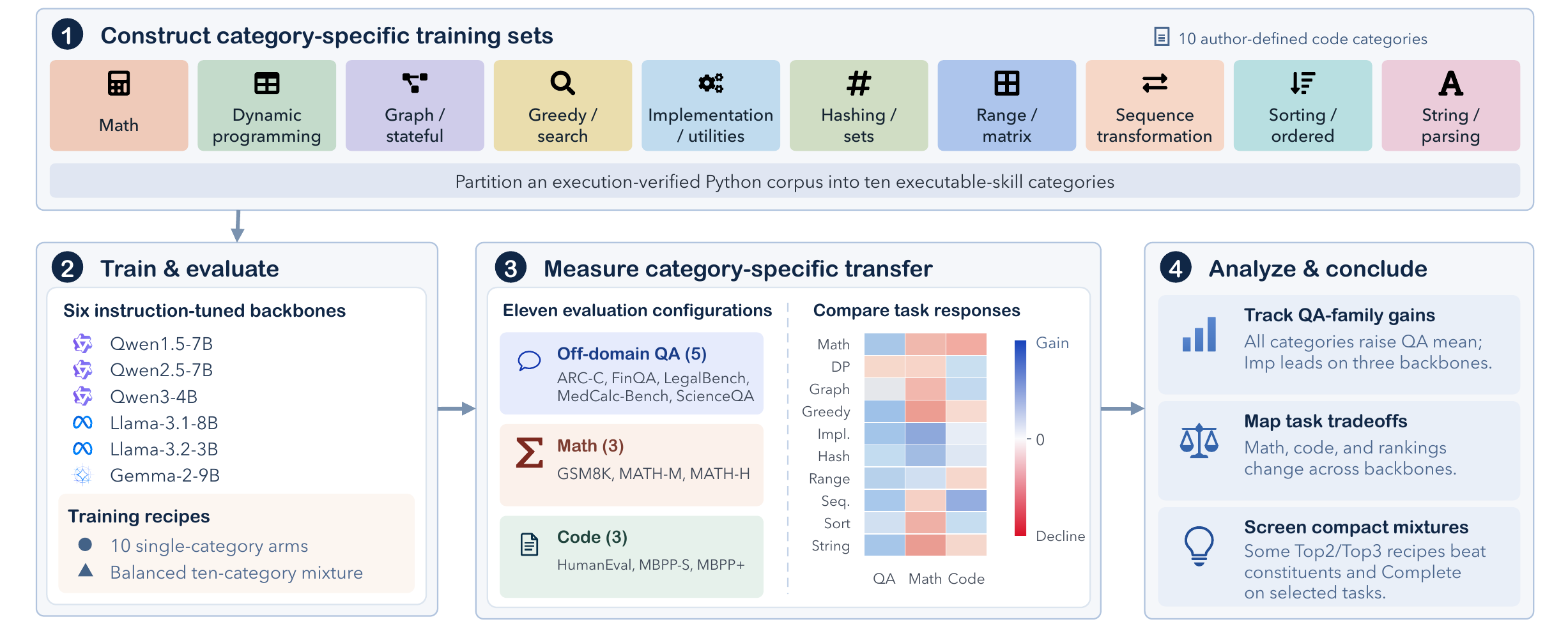}
\caption{\textbf{Overview of the category-specific transfer study.} Ten computational categories define single-category training configurations and an example-matched balanced mixture across six instruction-tuned backbones and eleven task configurations. The exploratory screen retains only mixtures whose constituents exceed Base and whose joint model exceeds both its best constituent and \textsc{Complete} on the same task. The heatmap is schematic.}
\label{fig:workflow}
\end{figure}

In this paper, we study this connection through a decompositional lens.
We first partition an execution-verified Python corpus derived from
KodCode \citep{xu2025kodcode} into ten interpretable categories based on
the dominant computational pattern of each solution. Across six instruction-tuned backbones and eleven evaluation configurations covering question answering, mathematics, and code generation, we evaluate single-category fine-tuning and balanced-mixture training relative to each backbone's starting checkpoint. We also compare categories with one another and with the example-matched balanced mixture to examine how their relative value varies across models and tasks. We then compose compact two- and three-category mixtures guided by these results and compare them with their individual constituents and full-corpus training.
Figure~\ref{fig:workflow} summarizes the workflow.

Our main results are summarized as follows:
\begin{itemize}[leftmargin=*]
\item \textbf{RQ1: Code data supports cross-task transfer, with
task-specific tradeoffs.}
The balanced mixture improves average question-answering performance
on all six backbones, while its effects on mathematics and code
generation vary across models. Positive task-family averages can
coexist with declines on individual tasks
(Section~\ref{sec:rq1}).

\item \textbf{RQ2: The best-performing category depends on both the
starting model and the target task.}
Category rankings can reverse when either the model or the task changes.
For example, on the medium-difficulty MATH configuration,
implementation/utilities outperforms dynamic programming for Llama-3.1,
while their ordering reverses for Llama-3.2.
For every backbone, at least one single-category model achieves a
higher average score across our evaluation suite than the
example-matched balanced mixture. These results show that broad
category coverage alone does not ensure stronger performance
(Section~\ref{sec:rq2}).

\item \textbf{RQ3: Compact mixtures can outperform both their best
constituent and full-corpus training.}
On selected model--task pairs spanning question answering, mathematics,
and code generation, two- and three-category mixtures outperform both
references while using roughly 10--15\% of the full corpus.
In these cases, each constituent also improves on the starting model
when trained individually.
These exploratory findings illustrate a \emph{less is more} pattern,
with successful compositions depending on the starting model and
target task
(Section~\ref{sec:rq3}).

\end{itemize}

\FloatBarrier

\section{Related Work}
\textbf{Code supervision.}
\citet{ma2024trainingstage} compare code exposure across training stages.
\citet{aryabumi2024tocode} study code proportions and quality during
pre-training, while \citet{zhang2025codingdata} examine coding-data
proportions during instruction tuning.
\citet{petty2025codepretraining} find gains in mathematics and structured
compositional tasks alongside losses on other linguistic and
knowledge-based tasks.
These findings motivate examining which code examples benefit a given
starting model and downstream task.
\citet{li2025codeio} introduce CodeI/O, which constructs input--output
prediction tasks with natural-language reasoning traces.
\citet{lin2025caco} propose Caco to synthesize and verify reasoning
supervision through executable code.
At inference time, \citet{gao2023pal} and \citet{chen2023pot}
delegate computation to interpreters in PAL and Program of Thoughts,
respectively.
Our study uses existing prompt--solution pairs for supervised post-training
of instruction-tuned models; downstream QA uses no execution tools.
Related studies examine how properties of code data affect transfer.
\citet{waheed2025codeinduced} investigate programming-language
representations and structural or semantic perturbations, while
\citet{twist2026not} study structural complexity.
Using influence functions, \citet{kou2025dataattributes} analyze how example
difficulty, reasoning behaviors, and token-level features contribute to
mathematical and code reasoning.
Complementing these analyses, we directly compare single-category and
joint training on groups defined by reference solutions' computational
patterns, such as dynamic programming and string parsing.

\textbf{Data selection and composition.}
\citet{ilyas2022datamodels} introduce Datamodels to relate training-subset
membership to model predictions.
\citet{xia2024less} propose LESS to select instruction-tuning examples
for target capabilities using gradient-based influence estimates.
\citet{liu2024deita} combine complexity, quality, and diversity measures
in DEITA, while \citet{lu2024instag} introduce InsTag to guide sampling
using semantic and intention tags.
Our categories describe how reference solutions compute answers,
complementing instruction-level tags.
For code generation, \citet{lv2025dataefficientcode} select fine-tuning
subsets based on complexity and alignment with the source-data distribution,
showing that selected subsets can outperform full-data training.
At the mixture level, \citet{xie2023doremi} introduce DoReMi to learn
pre-training domain weights using a proxy model.
\citet{liu2025regmix} and \citet{ye2025datamixinglaws} predict mixture
performance from smaller-scale experiments through RegMix and Data Mixing
Laws, respectively.
\citet{chen2023skillit} use skill dependencies to guide sampling in Skill-it,
while \citet{shin2026dynamixsft} adapt sampling across instruction-tuning
datasets through DynamixSFT.
\citet{dong2024datacomposition} examine benefits and conflicts from mixing
mathematical, code, and general instruction data.
We study composition within a code corpus, first comparing category-specific
transfer across models and tasks, then evaluating compact mixtures against
their individual constituents and full-corpus training.
We examine whether gains from individual categories persist or increase
when those categories are trained together.

\section{Study Design: From Code Categories to Compact Mixtures}
\label{sec:lens}
We study code supervision by separating examples according to their computational patterns, measuring each category's effects across models and tasks, and recombining categories into compact mixtures.
This connects overall transfer (RQ1), differences between categories (RQ2), and the effects of joint training (RQ3).

\subsection{Decomposing code by computational pattern}
\label{sec:categories}
We classify each prompt--solution pair by the dominant computational pattern of its reference solution: how the solution computes the answer. This author-defined taxonomy includes algorithmic strategies, such as \catlabel{\catDP}{recurrence} and \catlabel{\catGS}{greedy search}, and data-processing patterns, such as \catlabel{\catHash}{hash-based counting} and \catlabel{\catStr}{string parsing}. Table~\ref{tab:taxonomy} defines the ten categories through representative code patterns. Classification therefore concerns solution structure rather than the subject matter of the prompt.

The ten categories distinguish computational patterns while retaining enough examples for comparable training subsets.
Each example receives one primary category, yielding ten non-overlapping groups.
Auxiliary tags record additional algorithms or data structures but do not affect group assignment or training-subset selection.
Computational patterns can overlap across categories, so the groups do not isolate independent reasoning abilities.

We obtain primary-category labels in two AI-assisted stages.
First, an AI annotator assigns a category label and a confidence indicator.
An AI reviewer then examines low-confidence cases, cases with alternative-label suggestions, and a stratified sample across categories.
The reviewer sees the initial label; this AI review is not blind and does not provide independent human annotation.

\begin{table}[!htbp]
\caption{\textbf{Ten computational categories and their source-pool distribution.} Labels follow the dominant computational pattern (Section~\ref{sec:categories}); $n$ counts examples in each category before sampling. \catlabel{\catMath}{NT} denotes \catlabel{\catMath}{number theory}; \catlabel{\catIS}{impl.}\ denotes \catlabel{\catIS}{implementation}. Icons are reused in Table~\ref{tab:six_model_results} and the response maps.}
\label{tab:taxonomy}
\centering
\begin{minipage}[c]{0.68\textwidth}
\fontsize{7.0}{8.3}\selectfont
\setlength{\tabcolsep}{3pt}
\renewcommand{\arraystretch}{1}
\arrayrulecolor{tabrule}
\newcommand{\taxrow}{\rule[-5pt]{0pt}{17pt}}
\renewcommand{\pill}[1]{%
  \tikz[baseline=(pilltext.base)]{%
    \node[draw=pillline,fill=pillfill,rounded corners=2.5pt,%
          inner xsep=2.4pt,inner ysep=1.4pt,line width=0.3pt,%
          font=\fontsize{7.0}{8.3}\selectfont] (pilltext) {#1};}%
}
\renewcommand{\pillsep}{\hspace{2pt}}

\begin{tabularx}{\linewidth}{@{}lYr@{}}
\toprule
\textbf{Category} & \textbf{Representative patterns} & $\boldsymbol{n}$ \\
\midrule
\taxrow \catMath~ Math / NT & \mbox{\pill{Modular arith.}\pillsep\pill{Counting}\pillsep\pill{Transforms}} & 2,075 \\
\taxrow \catDP~ Dynamic prog. & \mbox{\pill{Recurrence}\pillsep\pill{Memoization}\pillsep\pill{Tables}} & 3,562 \\
\taxrow \catGR~ Graph / stateful & \mbox{\pill{Traversal}\pillsep\pill{Connectivity}\pillsep\pill{Paths}\pillsep\pill{States}} & 2,775 \\
\taxrow \catGS~ Greedy / search & \mbox{\pill{Greedy}\pillsep\pill{Binary/local search}\pillsep\pill{Rules}} & 1,706 \\
\taxrow \catIS~ Impl. / utilities & \mbox{\pill{Direct impl.}\pillsep\pill{Bookkeeping}\pillsep\pill{I/O glue}} & 5,621 \\
\taxrow \catHash~ Hashing / sets & \mbox{\pill{Hash maps}\pillsep\pill{Counting}\pillsep\pill{Sets}\pillsep\pill{Dedup}} & 5,077 \\
\taxrow \catRange~ Range / matrix & \mbox{\pill{Windows}\pillsep\pill{Prefix sums}\pillsep\pill{Intervals}\pillsep\pill{Grids}} & 4,221 \\
\taxrow \catSeq~ Sequence trans. & \mbox{\pill{Map/filter}\pillsep\pill{Encoding}\pillsep\pill{In-place ops}} & 4,440 \\
\taxrow \catSort~ Sorting / ordered & \mbox{\pill{Sorting}\pillsep\pill{Ranks}\pillsep\pill{Comparators}\pillsep\pill{Merge}} & 2,510 \\
\taxrow \catStr~ String / parsing & \mbox{\pill{Strings}\pillsep\pill{Tokenization}\pillsep\pill{Parsing}\pillsep\pill{Match}} & 3,987 \\
\bottomrule
\end{tabularx}
\end{minipage}\hfill
\begin{minipage}[c]{0.30\textwidth}
\centering
\includegraphics[width=\linewidth]{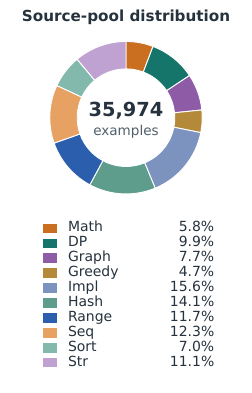}
\end{minipage}
\end{table}

\subsection{Measuring category-specific transfer}
For a given backbone and task, we measure transfer as the score change from the original checkpoint (\textsc{Base}):
\begin{equation}
\Delta = S_{\mathrm{trained}} - S_{\mathrm{Base}}.
\label{eq:response}
\end{equation}
Scores are expressed as percentages, so $\Delta$ is measured in percentage points (pp).
A positive value indicates improvement, and a negative value indicates decline.
For RQ1, we compare a balanced mixture of all ten categories with \textsc{Base} to assess transfer across task families.
For RQ2, we compare individual categories with one another and with the balanced mixture, then examine how rankings change across models and tasks.

\subsection{Recombining categories into compact mixtures}
\label{sec:mixture_comparison}
For RQ3, we combine selected categories into compact mixtures to test whether joint training preserves and extends the benefits observed under individual-category training.
We compare each mixture with its constituents trained individually and with full-corpus training (\textsc{Complete}).

We focus on cases in which joint training builds on categories that already improve the target task individually.
For each backbone and task, we therefore identify mixtures that meet both conditions:
\begin{enumerate}[nosep,leftmargin=*]
\item Each constituent category, when trained individually, scores higher than \textsc{Base}.
\item The jointly trained mixture scores higher than both its best constituent and \textsc{Complete}.
\end{enumerate}
The screen is exploratory: category ranking and mixture evaluation use the same evaluation suite, and qualifying cases are identified after observing the results.
Appendix~\ref{app:mixture_screening} reports the complete screening results and associated task scores.

Section~\ref{sec:setup} specifies the models, training recipes, and evaluation protocol.

\section{Experimental Setup}
\label{sec:setup}

\paragraph{Backbone models.}
We use six instruction-tuned backbone models: Qwen2.5-7B-Instruct, Qwen3-4B-Instruct-2507, Llama-3.1-8B-Instruct, Gemma-2-9B-Instruct, Llama-3.2-3B-Instruct, and Qwen1.5-7B-Chat.
All training runs start from the corresponding \textsc{Base} checkpoint.

\paragraph{Training.}
For RQ1 and RQ2, we apply LoRA supervised fine-tuning \citep{hu2022lora} separately to each computational category and to a balanced mixture of all ten categories.
Each single-category configuration and this balanced reference use 1,706 examples and 107 optimization steps.
Example counts and updates are matched across these trained configurations, but token counts are not.
Appendix~\ref{app:six_model_records} provides the sampling procedures and training hyperparameters.

\paragraph{Compact mixtures.}
We rank each backbone's single-category configurations by Overall, the equally weighted mean score across the eleven evaluation configurations described below.
Personalized Top2 and Top3 use the two or three highest-ranked categories, respectively.
Shared Top3 uses the same \catlabel{\catIS}{implementation/utilities (Imp)}, \catlabel{\catHash}{hashing/sets}, and \catlabel{\catSort}{sorting/ordered processing} combination for all six backbones.
Recipes are fixed before mixture training; their outcomes are then examined separately for each task using the criteria in Section~\ref{sec:mixture_comparison}.

Both Top3 recipes and their balanced and proportional all-category controls use 5,118 examples and 320 optimization steps.
This 5,118-example balanced control is separate from the 1,706-example balanced reference used for RQ1 and RQ2.
Top2 uses 3,412 examples and 214 optimization steps, with no corresponding equal-budget control.
\textsc{Complete} is a fresh LoRA fine-tuning run on the entire 35,974-example source pool.
Constituent, mixture, and \textsc{Complete} runs differ in example count and token exposure, so these comparisons do not isolate gains caused by combining categories.
Appendix~\ref{app:selection_protocol} provides the mixture budgets, sampling procedures, and training settings.

\paragraph{Evaluation.}
We evaluate eleven configurations spanning code generation, mathematics, and question answering (QA).
Code generation includes HumanEval \citep{chen2021humaneval}, MBPP-Simple, and MBPP+ \citep{austin2021mbpp,liu2023evalplus}.
Mathematics includes GSM8K \citep{cobbe2021gsm8k} and the medium- and high-level MATH configurations (MATH-M and MATH-H) \citep{hendrycks2021math}.
QA includes ARC-Challenge \citep{clark2018arc}, FinQA \citep{chen2021finqa}, LegalBench \citep{guha2023legalbench}, MedCalc-Bench \citep{khandekar2024medcalcbench}, and ScienceQA \citep{lu2022scienceqa}.
We use pass@1 for code generation and accuracy for the remaining tasks.
We report three-seed mean scores and their base-relative changes as defined in Equation~\ref{eq:response}.
Task-family averages weight their constituent configurations equally; Overall uses the eleven-configuration average defined above.
Table~\ref{tab:six_model_results} groups the five QA configurations separately from the six code and mathematics configurations; task-level analyses further separate code and mathematics.
Comparisons of these means are descriptive; we do not perform significance tests.
Appendix~\ref{app:six_model_records} provides the evaluation settings and record availability, while Appendix~\ref{app:six_model_scores} presents the complete single-category and balanced-mixture results.

\section{RQ1: Does code post-training improve performance across tasks and models?}
\label{sec:results}
\label{sec:rq1}
We compare each backbone's starting checkpoint (Base) with the same model after balanced-mixture fine-tuning on all ten code categories. This comparison tests whether code-only post-training transfers across task families.

\begin{finding}
\textbf{Answer.} The balanced mixture improves the QA mean on all six models, but code and mathematics gains depend on the backbone. Positive family means also conceal task-level declines.
\end{finding}

\begin{figure}[!t]
\centering
\includegraphics[width=0.90\textwidth,trim=0 19bp 0 3bp,clip]{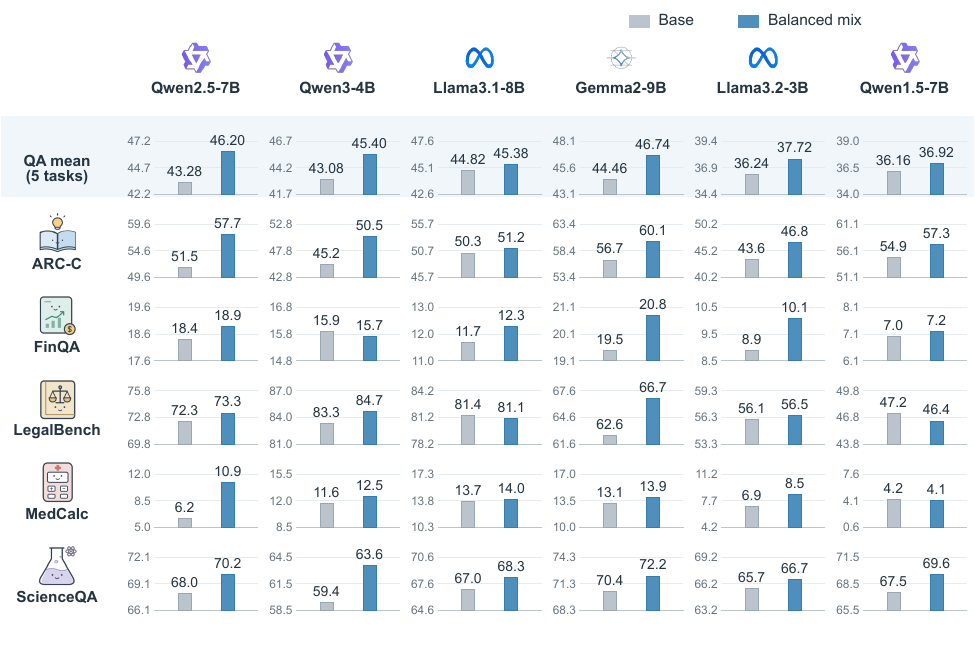}
\caption{\textbf{QA gains are consistent across models but uneven across tasks.} Bars compare Base and Balanced mix (three-seed means, \%) for the five-task QA mean and each task. QA means use equal weights and released one-decimal task scores. Each row uses the same numerical span across models, with panel-specific truncated baselines and absolute-score ticks.}
\label{fig:rq1_qa_tasks}
\end{figure}

\paragraph{QA improves across all six backbones.}
The balanced mixture raises the five-task QA mean by $+0.6$ to $+2.9$ pp (Figure~\ref{fig:rq1_qa_tasks}). It improves 26 of the 30 model--task scores, including ARC-Challenge and ScienceQA on every backbone. The four declines occur on Qwen3 FinQA, Llama-3.1 LegalBench, and Qwen1.5 LegalBench and MedCalc. Thus, the positive QA mean reflects broad transfer while retaining exceptions that an aggregate alone would hide.

\begin{figure}[!t]
\centering
\includegraphics[width=0.90\textwidth,trim=0 15bp 0 3bp,clip]{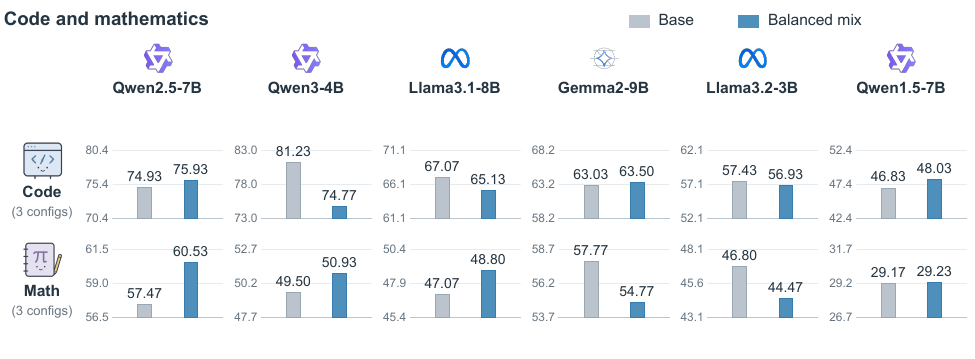}
\caption{\textbf{Code and mathematics gains depend on the backbone.} Bars average HumanEval, MBPP+, and MBPP-Simple (Code), or GSM8K, MATH-M, and MATH-H (Math), with equal weights. Reporting conventions follow Figure~\ref{fig:rq1_qa_tasks}.}
\label{fig:rq1_code_math}
\end{figure}

\paragraph{Code and mathematics show different model-dependent responses.}
The balanced mixture improves the Code mean on three backbones and the Math mean on four, including Qwen1.5's small $+0.1$ pp Math gain (Figure~\ref{fig:rq1_code_math}). Qwen3 gains $+1.4$ pp in Math while losing $6.5$ pp in Code. Gemma-2 instead gains $+0.5$ pp in Code while losing $3.0$ pp in Math. Llama-3.2 declines in both families, despite its positive QA mean. Separating Code and Math therefore exposes responses that their combined mean would obscure.

\paragraph{QA gains extend across individual code categories.}
The sixty single-category training configurations provide complementary context in Appendix Table~\ref{tab:six_model_results}. Using their reported base-relative deltas, all sixty improve the QA mean and 44 improve Overall. Only 24 improve the combined code-and-mathematics mean, reinforcing the uneven transfer pattern. A separate single-seed audit checks how the positive QA means depend on task composition (Appendix~\ref{app:finding1_audit}). All sixty configurations retain positive five-task QA means in this audit, but only 45 remain positive after excluding ARC-Challenge and ScienceQA. QA mean gains therefore extend across categories and backbones, although their distribution across individual QA tasks remains uneven.
\section{RQ2: Which code categories help which tasks on which models?}
\label{sec:rq2}
\begin{figure}[!t]
\centering
\includegraphics[width=0.95\textwidth,trim=0 15bp 0 3bp,clip]{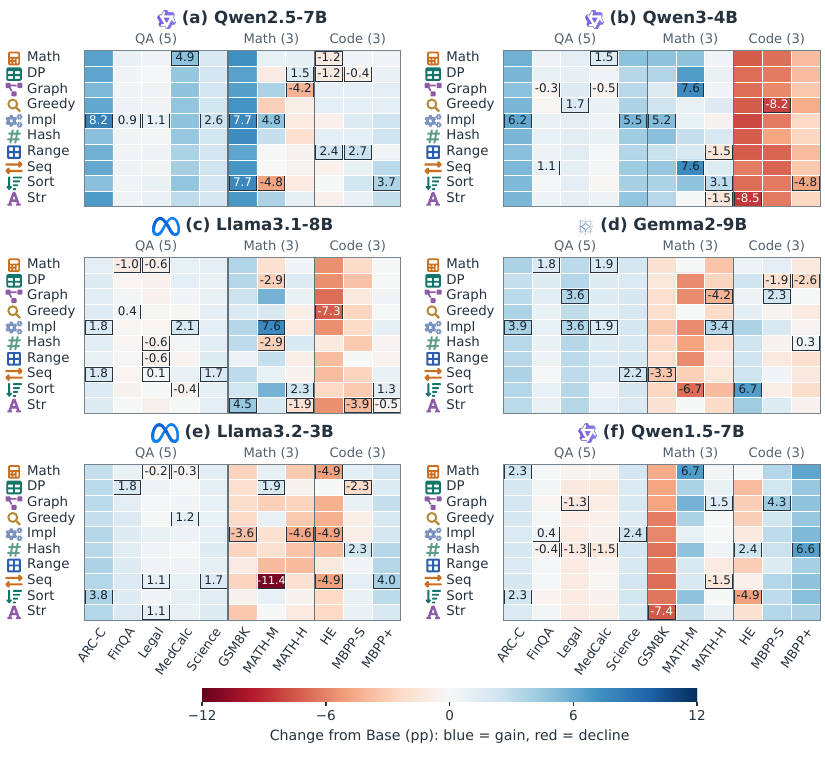}
\caption{\textbf{Backbone-level category--task responses.} All $6\times10\times11=660$ cells show three-seed mean changes from the corresponding Base (pp). Rows are categories; columns group QA, math, and code tasks. Comparing rows within a task reveals category preferences; comparing panels reveals their dependence on the backbone. Blue/red denotes gain/decline on a shared scale. Outlined numbers mark the largest positive and negative response per task within each backbone, retaining ties; these are descriptive extrema. Appendix~\ref{app:six_model_scores} gives every value and the balanced-mixture rows.}
\label{fig:six_task_map}
\end{figure}

We now compare code categories with one another while holding the backbone, target task, example count, and optimization steps fixed. Figure~\ref{fig:six_task_map} shows all $6\times10\times11=660$ category--task responses, allowing both task-dependent and model-dependent category preferences to be examined.

\begin{finding}
\textbf{Answer.} A code category's value depends on both the target task and the starting model. Recurring strengths exist, but no category is the preferred choice across all objectives and backbones.
\end{finding}

\paragraph{Changing the target task can reverse the category ranking.}
On Llama-3.1, \catlabel{\catIS}{implementation (Imp)} improves MATH-M by $+7.6$ pp, whereas \catlabel{\catDP}{dynamic programming} lowers it by $2.9$ pp. On MATH-H, \catlabel{\catIS}{Imp} instead changes the score by $-1.1$ pp and \catlabel{\catDP}{dynamic programming} improves it by $+1.5$ pp. The category ordering therefore reverses between these two mathematics configurations. A category's downstream value depends on target tasks, even within the same task family.

\paragraph{Changing the backbone model also changes the preferred category.}
Holding MATH-M fixed, the same \catlabel{\catIS}{Imp}--\catlabel{\catDP}{dynamic-programming} comparison reverses between the two Llama backbones. \catlabel{\catIS}{Imp} leads on Llama-3.1, but on Llama-3.2 it lowers the score by $1.9$ pp while \catlabel{\catDP}{dynamic programming} improves it by $+1.9$ pp. These reversals show that the same target task can favor different categories on different starting models.

\paragraph{Aggregate rankings also depend on the backbone.}
For example, \catlabel{\catIS}{Imp} leads Overall on Qwen2.5, whereas \catlabel{\catSeq}{sequence transformations} leads on Qwen3 and \catlabel{\catSort}{sorting/ordered processing} on Llama-3.2. Each backbone has a single-category Overall score above its example-matched balanced mixture. The largest margin per backbone ranges from $+0.5$ to $+1.2$ pp (Appendix~\ref{app:aggregate_results}). Thus, balanced sampling across all ten categories does not yield the highest observed Overall score under this example budget.

\paragraph{A recurring strength still has task-specific limits.}
\catlabel{\catIS}{Imp} raises the QA-family mean on all six backbones and gives the largest single-category QA gain on Qwen2.5, Qwen3, and Llama-3.1. Its respective gains on these three models are $3.5$, $2.9$, and $0.9$ pp. Yet \catlabel{\catIS}{Imp} lowers FinQA and LegalBench by $0.2$ pp each on Llama-3.1, and LegalBench and MedCalc by $1.1$ and $0.5$ pp on Qwen1.5. A recurring advantage in the QA mean therefore coexists with declines on specific tasks.

\FloatBarrier

\section{RQ3: Can compact mixtures outperform single-category and full-corpus training?}
\label{sec:compact_mixture_results}
\label{sec:rq3}
We next examine joint training of the Overall-ranked personalized recipes and the shared Top3 recipe defined in Section~\ref{sec:setup}. After training, we compare their outcomes separately on each evaluation configuration, using the individual constituents and full-corpus training (\textsc{Complete}) as references.

\begin{finding}
\textbf{Answer.} Compact mixtures can outperform both their best constituent and full-corpus training while using only 9.5--14.2\% of the examples. These gains occur on selected model--task pairs whose constituent categories each improve on Base.
\end{finding}

\begin{figure}[!htbp]
\centering
\includegraphics[width=\textwidth]{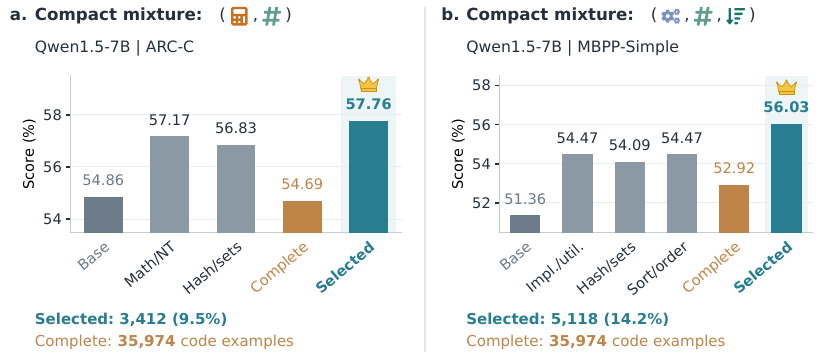}
\caption{\textbf{Two qualifying compact mixtures on Qwen1.5.} Panel (a) shows \catlabel{\catMath}{Math} + \catlabel{\catHash}{Hash} on ARC-C; panel (b) shows \catlabel{\catIS}{Imp} + \catlabel{\catHash}{Hash} + \catlabel{\catSort}{Sort} on MBPP-Simple. Selected denotes the jointly trained mixture, and Complete denotes training on all 35,974 examples. Bars show the means on truncated y axes; yellow crowns mark the highest mean. Counts and percentages give training-set sizes and fractions of Complete. See Appendix~\ref{app:mixture_screening} for all screened examples.}
\label{fig:compact_mixture}
\end{figure}

\paragraph{Coverage of the task-level screen.}
We apply the conditions in Section~\ref{sec:mixture_comparison} to the trained recipes. Each comparison corresponds to one mixture, one backbone, and one evaluation configuration. The screen retains \textbf{9 of 66 two-category comparisons} and \textbf{18 of 132 three-category comparisons}. The denominators reflect one Top2 recipe and two Top3 recipes per backbone, each evaluated on eleven configurations across six backbones. Together, the 27 qualifying comparisons span five backbones and nine evaluation configurations. Figure~\ref{fig:compact_mixture} illustrates one qualifying recipe of each size; Appendix~\ref{app:mixture_screening} reports the complete screen.

\paragraph{Representative gains over both references.}
On Qwen1.5 ARC-Challenge, the \catlabel{\catMath}{Math}/\catlabel{\catHash}{Hash} Top2 mixture reaches $57.76\%$, exceeding its best constituent by $0.60$ pp and \textsc{Complete} by $3.07$ pp. It uses 3,412 examples, or 9.5\% of the full corpus. On the same backbone's MBPP-Simple, the shared \catlabel{\catIS}{Imp}/\catlabel{\catHash}{Hash}/\catlabel{\catSort}{Sort} Top3 reaches $56.03\%$, exceeding its best constituent by $1.56$ pp and \textsc{Complete} by $3.11$ pp. It uses 5,118 examples, or 14.2\% of the full corpus. Every constituent in both examples individually outperforms Base; Figure~\ref{fig:compact_mixture} shows the corresponding absolute scores.

\paragraph{Successful combinations vary across models and tasks.}
Even for the same target task, different model-specific recipes satisfy the screen. On ARC-Challenge, the qualifying Top2 recipe combines \catlabel{\catIS}{Imp}/\catlabel{\catStr}{Str} on Qwen2.5 and \catlabel{\catSort}{Sort}/\catlabel{\catIS}{Imp} on Llama-3.1. These examples reflect task-level outcomes of recipes selected by Overall ranking, rather than separate category selection for ARC-Challenge. The matched-budget Top3 comparisons also show that combining individually high-ranked categories does not consistently outperform balanced sampling on Overall (Appendix~\ref{app:mixture_overall}). Individual-category rankings therefore provide candidates for joint training, but do not directly determine which mixtures will perform best. Appendix~\ref{app:selection_protocol} gives the full Top3 task maps, matched-budget controls, and generation-budget sensitivity.
\section{Discussion}
\label{sec:discussion}
\textbf{Category selection and mixture evaluation answer different questions.} Category-level evaluation can narrow the candidate training pool, but single-category rankings do not determine performance after joint training. The compact-mixture results and matched-budget comparisons show why both stages matter when selecting code data. Likewise, improvements in Overall or task-family means do not establish gains on a particular target task. Code data selection therefore benefits from examining both the composition of the training set and the distribution of gains across the intended tasks.

\textbf{The evidence is specific to Python and the present taxonomy.} Our training pool contains execution-verified Python solutions, so the observed category rankings and task responses may not extend to other languages. The ten-category taxonomy records each solution's dominant computational pattern. Broad groups such as \catlabel{\catIS}{implementation/utilities} may contain finer distinctions that this analysis does not resolve. We therefore interpret category responses as effects of the defined training subsets, rather than effects of isolated reasoning abilities.

\textbf{The mixture screen establishes selected cases, not a general selection rule.} The space of possible mixtures extends beyond the recipes evaluated here. We tested one Personalized Top2, one Personalized Top3, and one Shared Top3 recipe per backbone (Appendix~\ref{app:selection_protocol}). Personalized recipes use Overall rankings from the same evaluation suite, and qualifying task-level cases are identified after training. Moreover, constituent, mixture, and \textsc{Complete} runs differ in training exposure. The observed advantages therefore establish neither an optimal composition nor causal synergy between categories. Independent evaluation of frozen recipes against matched-budget controls would be needed to assess a generalizable selection rule.

\textbf{The mechanisms behind category-dependent transfer remain unresolved.} The single-seed audit of \catlabel{\catIS}{Imp} contains examples compatible with improved application of rules and stated relationships (Appendix~\ref{app:imp_qa_audit}). However, QA gains and reversals in mathematics and code may also reflect changes in output behavior or evaluation interfaces. The item-level audits in Appendices~\ref{app:finding1_audit} and~\ref{app:imp_qa_audit} do not establish a causal mechanism or quantify variability across training seeds. They identify candidate explanations for the observed transfer patterns, whose causal contributions remain to be tested.

\section{Conclusion}
Our work introduces a decompositional lens for studying how computational
categories of code data affect LLM post-training. By examining categories
individually and in combination, we connect the analysis of transfer with
the construction of compact training mixtures. Our experiments reveal
recurring gains in average question-answering performance, alongside
task-level tradeoffs and category preferences that vary across models
and evaluation objectives. These findings highlight the importance of
understanding which categories benefit the particular model and task
being considered.

The mixture experiments further demonstrate the potential of category
composition. On selected model--task pairs, compact mixtures of individually
beneficial categories outperform both their best constituent and
full-corpus training while using roughly 10--15\% of the full corpus.
These exploratory findings illustrate a \emph{less is more} pattern and
motivate strategies for selecting and combining code data that account
for both category-specific transfer and the effects of joint training.
\label{sec:main_end}
\FloatBarrier
\clearpage
\linespread{1}\selectfont
\bibliographystyle{iclr2027_conference}
\bibliography{reference}

@inproceedings{zhang2025codingdata,
  title = {Unveiling the Impact of Coding Data Instruction Fine-Tuning on Large Language Models Reasoning},
  author = {Zhang, Xinlu and Chen, Zhiyu Zoey and Ye, Xi and Yang, Xianjun and Chen, Lichang and Wang, William Yang and Petzold, Linda Ruth},
  booktitle = {Proceedings of the AAAI Conference on Artificial Intelligence},
  volume = {39},
  pages = {25949--25957},
  year = {2025},
  doi = {10.1609/aaai.v39i24.34789},
  url = {https://ojs.aaai.org/index.php/AAAI/article/view/34789}
}

@article{cobbe2021gsm8k,
  title = {Training Verifiers to Solve Math Word Problems},
  author = {Cobbe, Karl and Kosaraju, Vineet and Bavarian, Mohammad and Chen, Mark and Jun, Heewoo and Kaiser, Lukasz and Plappert, Matthias and Tworek, Jerry and Hilton, Jacob and Nakano, Reiichiro and Hesse, Christopher and Schulman, John},
  journal = {arXiv preprint arXiv:2110.14168},
  year = {2021},
  url = {https://arxiv.org/abs/2110.14168}
}

@inproceedings{ma2024trainingstage,
  title = {{At Which Training Stage Does Code Data Help LLMs Reasoning?}},
  author = {Ma, Yingwei and Liu, Yue and Yu, Yue and Zhang, Yuanliang and Jiang, Yu and Wang, Changjian and Li, Shanshan},
  booktitle = {The Twelfth International Conference on Learning Representations},
  year = {2024},
  url = {https://proceedings.iclr.cc/paper_files/paper/2024/hash/9c2aa1e456ea543997f6927295196381-Abstract-Conference.html}
}

@article{petty2025codepretraining,
  title = {{How Does Code Pretraining Affect Language Model Task Performance?}},
  author = {Petty, Jackson and van Steenkiste, Sjoerd and Linzen, Tal},
  journal = {Transactions on Machine Learning Research},
  year = {2025},
  url = {https://openreview.net/forum?id=pxxmUKKgel}
}

@inproceedings{li2025codeio,
  title = {{CodeIO: Condensing Reasoning Patterns via Code Input-Output Prediction}},
  author = {Li, Junlong and Guo, Daya and Yang, Dejian and Xu, Runxin and Wu, Yu and He, Junxian},
  booktitle = {Proceedings of the 42nd International Conference on Machine Learning},
  volume = {267},
  pages = {34471--34489},
  year = {2025},
  series = {Proceedings of Machine Learning Research},
  publisher = {PMLR},
  url = {https://proceedings.mlr.press/v267/li25t.html}
}

@inproceedings{lin2025caco,
  title = {{Scaling Code-Assisted Chain-of-Thoughts and Instructions for Model Reasoning}},
  author = {Lin, Honglin and Pei, Qizhi and Pan, Zhuoshi and Li, Yu and Gao, Xin and Li, Juntao and He, Conghui and Wu, Lijun},
  booktitle = {Advances in Neural Information Processing Systems},
  volume = {38},
  pages = {35204--35237},
  year = {2025},
  publisher = {Curran Associates, Inc.},
  doi = {10.52202/085713-1182},
  url = {https://proceedings.neurips.cc/paper_files/paper/2025/hash/3267aa172c31ae3641c32ecc6e42e5cd-Abstract-Conference.html}
}

@inproceedings{kou2025dataattributes,
  title = {{Which Data Attributes Stimulate Math and Code Reasoning? An Investigation via Influence Functions}},
  author = {Kou, Siqi and Tian, Qingyuan and Xu, Hanwen and Zeng, Zihao and Deng, Zhijie},
  booktitle = {Advances in Neural Information Processing Systems},
  volume = {38},
  pages = {12302--12324},
  year = {2025},
  publisher = {Curran Associates, Inc.},
  doi = {10.52202/085713-0412},
  url = {https://proceedings.neurips.cc/paper_files/paper/2025/hash/12323a9ca6eb70fdc993efa13c207f88-Abstract-Conference.html}
}

@inproceedings{dong2024datacomposition,
  title = {{How Abilities in Large Language Models are Affected by Supervised Fine-tuning Data Composition}},
  author = {Dong, Guanting and Yuan, Hongyi and Lu, Keming and Li, Chengpeng and Xue, Mingfeng and Liu, Dayiheng and Wang, Wei and Yuan, Zheng and Zhou, Chang and Zhou, Jingren},
  booktitle = {Proceedings of the 62nd Annual Meeting of the Association for Computational Linguistics (Volume 1: Long Papers)},
  pages = {177--198},
  year = {2024},
  doi = {10.18653/v1/2024.acl-long.12},
  url = {https://aclanthology.org/2024.acl-long.12/}
}

@inproceedings{lv2025dataefficientcode,
  title = {{Data-efficient LLM Fine-tuning for Code Generation}},
  author = {Lv, Weijie and Xia, Xuan and Huang, Sheng-Jun},
  booktitle = {2025 International Joint Conference on Neural Networks (IJCNN)},
  pages = {1--8},
  year = {2025},
  doi = {10.1109/IJCNN64981.2025.11227906},
  url = {https://doi.org/10.1109/IJCNN64981.2025.11227906}
}

@inproceedings{liu2024deita,
  title = {{What Makes Good Data for Alignment? A Comprehensive Study of Automatic Data Selection in Instruction Tuning}},
  author = {Liu, Wei and Zeng, Weihao and He, Keqing and Jiang, Yong and He, Junxian},
  booktitle = {The Twelfth International Conference on Learning Representations},
  year = {2024},
  url = {https://proceedings.iclr.cc/paper_files/paper/2024/hash/6091f2bb355e960600f62566ac0e2862-Abstract-Conference.html}
}

@inproceedings{shin2026dynamixsft,
  title = {{DynamixSFT: Dynamic Mixture Optimization of Instruction Tuning Collections}},
  author = {Shin, Haebin and Ji, Lei and Liu, Xiao and Yu, Zhiwei and Yoo, Hyunwoo and Chen, Qi and Gong, Yeyun},
  booktitle = {Findings of the Association for Computational Linguistics: ACL 2026},
  pages = {39590--39603},
  year = {2026},
  doi = {10.18653/v1/2026.findings-acl.1972},
  url = {https://aclanthology.org/2026.findings-acl.1972/}
}

@inproceedings{liu2025regmix,
  title = {{RegMix: Data Mixture as Regression for Language Model Pre-training}},
  author = {Liu, Qian and Zheng, Xiaosen and Muennighoff, Niklas and Zeng, Guangtao and Dou, Longxu and Pang, Tianyu and Jiang, Jing and Lin, Min},
  booktitle = {The Thirteenth International Conference on Learning Representations},
  year = {2025},
  url = {https://proceedings.iclr.cc/paper_files/paper/2025/hash/5f67d864aae6115374fed7beddd119e0-Abstract-Conference.html}
}

@inproceedings{lu2024instag,
  title = {{\#InsTag: Instruction Tagging for Analyzing Supervised Fine-tuning of Large Language Models}},
  author = {Lu, Keming and Yuan, Hongyi and Yuan, Zheng and Lin, Runji and Lin, Junyang and Tan, Chuanqi and Zhou, Chang and Zhou, Jingren},
  booktitle = {The Twelfth International Conference on Learning Representations},
  year = {2024},
  url = {https://proceedings.iclr.cc/paper_files/paper/2024/hash/9dae2a90bae49dc874ce1ca8fcc20879-Abstract-Conference.html}
}

@inproceedings{ye2025datamixinglaws,
  title = {{Data Mixing Laws: Optimizing Data Mixtures by Predicting Language Modeling Performance}},
  author = {Ye, Jiasheng and Liu, Peiju and Sun, Tianxiang and Zhan, Jun and Zhou, Yunhua and Qiu, Xipeng},
  booktitle = {The Thirteenth International Conference on Learning Representations},
  year = {2025},
  url = {https://proceedings.iclr.cc/paper_files/paper/2025/hash/cc84bfabe6389d8883fc2071c848f62a-Abstract-Conference.html}
}

@inproceedings{xia2024less,
  title = {{LESS}: Selecting Influential Data for Targeted Instruction Tuning},
  author = {Xia, Mengzhou and Malladi, Sadhika and Gururangan, Suchin and Arora, Sanjeev and Chen, Danqi},
  booktitle = {Proceedings of the 41st International Conference on Machine Learning},
  volume = {235},
  pages = {54104--54132},
  year = {2024},
  series = {Proceedings of Machine Learning Research},
  publisher = {PMLR},
  url = {https://proceedings.mlr.press/v235/xia24c.html}
}

@article{twist2026not,
  title = {Not All Code Is Equal: A Data-Centric Study of Code Complexity and LLM Reasoning},
  author = {Twist, Lukas and Yang, Shu and Yan, Hanqi and Gong, Jingzhi and Wang, Di and Yannakoudakis, Helen and Zhang, Jie M},
  journal = {arXiv preprint arXiv:2601.21894},
  year = {2026},
  url = {https://arxiv.org/abs/2601.21894}
}

@article{hendrycks2021math,
  title = {Measuring Mathematical Problem Solving with the {MATH} Dataset},
  author = {Hendrycks, Dan and Burns, Collin and Kadavath, Saurav and Arora, Akul and Basart, Steven and Tang, Eric and Song, Dawn and Steinhardt, Jacob},
  journal = {arXiv preprint arXiv:2103.03874},
  year = {2021},
  url = {https://arxiv.org/abs/2103.03874}
}

@article{clark2018arc,
  title = {Think You Have Solved Question Answering? {T}ry {ARC}, the {AI2} Reasoning Challenge},
  author = {Clark, Peter and Cowhey, Isaac and Etzioni, Oren and Khot, Tushar and Sabharwal, Ashish and Schoenick, Carissa and Tafjord, Oyvind},
  journal = {arXiv preprint arXiv:1803.05457},
  year = {2018},
  url = {https://arxiv.org/abs/1803.05457}
}

@inproceedings{chen2021finqa,
  title = {{FinQA}: A Dataset of Numerical Reasoning over Financial Data},
  author = {Chen, Zhiyu and Chen, Wenhu and Smiley, Charese and Shah, Sameena and Borova, Iana and Langdon, Dylan and Moussa, Reema and Beane, Matt and Huang, Ting-Hao and Routledge, Bryan and Wang, William Yang},
  booktitle = {Proceedings of the 2021 Conference on Empirical Methods in Natural Language Processing},
  pages = {3697--3711},
  year = {2021},
  publisher = {Association for Computational Linguistics},
  doi = {10.18653/v1/2021.emnlp-main.300},
  url = {https://aclanthology.org/2021.emnlp-main.300/}
}

@inproceedings{guha2023legalbench,
  title = {{LegalBench}: A Collaboratively Built Benchmark for Measuring Legal Reasoning in Large Language Models},
  author = {Guha, Neel and Nyarko, Julian and Ho, Daniel E. and R\'{e}, Christopher and Chilton, Adam and Narayana, Aditya and Chohlas-Wood, Alex and Peters, Austin and Waldon, Brandon and Rockmore, Daniel N. and Zambrano, Diego and Talisman, Dmitry and Hoque, Enam and Surani, Faiz and Fagan, Frank and Sarfaty, Galit and Dickinson, Gregory M. and Porat, Haggai and Hegland, Jason and Wu, Jessica and Nudell, Joe and Niklaus, Joel and Nay, John and Choi, Jonathan H. and Tobia, Kevin and Hagan, Margaret and Ma, Megan and Livermore, Michael and Rasumov-Rahe, Nikon and Holzenberger, Nils and Kolt, Noam and Henderson, Peter and Rehaag, Sean and Goel, Sharad and Gao, Shang and Williams, Spencer and Gandhi, Sunny and Zur, Tom and Iyer, Varun and Li, Zehua},
  booktitle = {Advances in Neural Information Processing Systems},
  volume = {36},
  pages = {44123--44279},
  year = {2023},
  publisher = {Curran Associates, Inc.},
  doi = {10.52202/075280-1915},
  url = {https://proceedings.neurips.cc/paper_files/paper/2023/hash/89e44582fd28ddfea1ea4dcb0ebbf4b0-Abstract-Datasets_and_Benchmarks.html}
}

@article{waheed2025codeinduced,
  title = {On Code-Induced Reasoning in {LLM}s},
  author = {Waheed, Abdul and Wu, Zhen and Ros\'{e}, Carolyn and Ippolito, Daphne},
  journal = {arXiv preprint arXiv:2509.21499},
  year = {2025},
  url = {https://arxiv.org/abs/2509.21499}
}

@inproceedings{chen2023skillit,
  title = {Skill-it! {A} Data-Driven Skills Framework for Understanding and Training Language Models},
  author = {Chen, Mayee F. and Roberts, Nicholas and Bhatia, Kush and Wang, Jue and Zhang, Ce and Sala, Frederic and R\'{e}, Christopher},
  booktitle = {Advances in Neural Information Processing Systems},
  volume = {36},
  pages = {36000--36040},
  year = {2023},
  publisher = {Curran Associates, Inc.},
  doi = {10.52202/075280-1562},
  url = {https://proceedings.neurips.cc/paper_files/paper/2023/hash/70b8505ac79e3e131756f793cd80eb8d-Abstract-Conference.html}
}

@inproceedings{gao2023pal,
  title = {{PAL}: Program-aided Language Models},
  author = {Gao, Luyu and Madaan, Aman and Zhou, Shuyan and Alon, Uri and Liu, Pengfei and Yang, Yiming and Callan, Jamie and Neubig, Graham},
  booktitle = {Proceedings of the 40th International Conference on Machine Learning},
  volume = {202},
  pages = {10764--10799},
  year = {2023},
  series = {Proceedings of Machine Learning Research},
  publisher = {PMLR},
  url = {https://proceedings.mlr.press/v202/gao23f.html}
}

@article{chen2023pot,
  title = {Program of Thoughts Prompting: Disentangling Computation from Reasoning for Numerical Reasoning Tasks},
  author = {Chen, Wenhu and Ma, Xueguang and Wang, Xinyi and Cohen, William W.},
  journal = {Transactions on Machine Learning Research},
  year = {2023},
  url = {https://openreview.net/forum?id=YfZ4ZPt8zd}
}

@article{aryabumi2024tocode,
  title = {To Code, or Not To Code? {E}xploring Impact of Code in Pre-training},
  author = {Aryabumi, Viraat and Su, Yixuan and Ma, Raymond and Morisot, Adrien and Zhang, Ivan and Locatelli, Acyr and Fadaee, Marzieh and \"Ust\"un, Ahmet and Hooker, Sara},
  journal = {arXiv preprint arXiv:2408.10914},
  year = {2024},
  url = {https://arxiv.org/abs/2408.10914}
}

@inproceedings{hu2022lora,
  title = {{LoRA}: Low-Rank Adaptation of Large Language Models},
  author = {Hu, Edward J. and Shen, Yelong and Wallis, Phillip and Allen-Zhu, Zeyuan and Li, Yuanzhi and Wang, Shean and Wang, Lu and Chen, Weizhu},
  booktitle = {International Conference on Learning Representations},
  year = {2022},
  url = {https://openreview.net/forum?id=nZeVKeeFYf9}
}

@inproceedings{xu2025kodcode,
  title = {{KodCode}: A Diverse, Challenging, and Verifiable Synthetic Dataset for Coding},
  author = {Xu, Zhangchen and Liu, Yang and Yin, Yueqin and Zhou, Mingyuan and Poovendran, Radha},
  booktitle = {Findings of the Association for Computational Linguistics: ACL 2025},
  pages = {6980--7008},
  year = {2025},
  publisher = {Association for Computational Linguistics},
  doi = {10.18653/v1/2025.findings-acl.365},
  url = {https://aclanthology.org/2025.findings-acl.365/}
}

@inproceedings{xie2023doremi,
  title = {{DoReMi}: Optimizing Data Mixtures Speeds Up Language Model Pretraining},
  author = {Xie, Sang Michael and Pham, Hieu and Dong, Xuanyi and Du, Nan and Liu, Hanxiao and Lu, Yifeng and Liang, Percy S and Le, Quoc V and Ma, Tengyu and Yu, Adams Wei},
  booktitle = {Advances in Neural Information Processing Systems},
  volume = {36},
  pages = {69798--69818},
  year = {2023},
  doi = {10.52202/075280-3059},
  url = {https://proceedings.neurips.cc/paper_files/paper/2023/hash/dcba6be91359358c2355cd920da3fcbd-Abstract-Conference.html}
}

@inproceedings{ilyas2022datamodels,
  title = {Datamodels: Understanding Predictions with Data and Data with Predictions},
  author = {Ilyas, Andrew and Park, Sung Min and Engstrom, Logan and Leclerc, Guillaume and Madry, Aleksander},
  booktitle = {Proceedings of the 39th International Conference on Machine Learning},
  volume = {162},
  pages = {9525--9587},
  year = {2022},
  series = {Proceedings of Machine Learning Research},
  publisher = {PMLR},
  url = {https://proceedings.mlr.press/v162/ilyas22a.html}
}

@article{chen2021humaneval,
  title = {Evaluating Large Language Models Trained on Code},
  author = {Chen, Mark and Tworek, Jerry and Jun, Heewoo and Yuan, Qiming and Pinto, Henrique Ponde de Oliveira and Kaplan, Jared and Edwards, Harri and Burda, Yuri and Joseph, Nicholas and Brockman, Greg and Ray, Alex and Puri, Raul and Krueger, Gretchen and Petrov, Michael and Khlaaf, Heidy and Sastry, Girish and Mishkin, Pamela and Chan, Brooke and Gray, Scott and Ryder, Nick and Pavlov, Mikhail and Power, Alethea and Kaiser, Lukasz and Bavarian, Mohammad and Winter, Clemens and Tillet, Philippe and Such, Felipe Petroski and Cummings, Dave and Plappert, Matthias and Chantzis, Fotios and Barnes, Elizabeth and Herbert-Voss, Ariel and Guss, William Hebgen and Nichol, Alex and Paino, Alex and Tezak, Nikolas and Tang, Jie and Babuschkin, Igor and Balaji, Suchir and Jain, Shantanu and Saunders, William and Hesse, Christopher and Carr, Andrew N. and Leike, Jan and Achiam, Josh and Misra, Vedant and Morikawa, Evan and Radford, Alec and Knight, Matthew and Brundage, Miles and Murati, Mira and Mayer, Katie and Welinder, Peter and McGrew, Bob and Amodei, Dario and McCandlish, Sam and Sutskever, Ilya and Zaremba, Wojciech},
  journal = {arXiv preprint arXiv:2107.03374},
  year = {2021},
  doi = {10.48550/arXiv.2107.03374},
  url = {https://arxiv.org/abs/2107.03374}
}

@article{austin2021mbpp,
  title = {Program Synthesis with Large Language Models},
  author = {Austin, Jacob and Odena, Augustus and Nye, Maxwell and Bosma, Maarten and Michalewski, Henryk and Dohan, David and Jiang, Ellen and Cai, Carrie and Terry, Michael and Le, Quoc and Sutton, Charles},
  journal = {arXiv preprint arXiv:2108.07732},
  year = {2021},
  doi = {10.48550/arXiv.2108.07732},
  url = {https://arxiv.org/abs/2108.07732}
}

@inproceedings{liu2023evalplus,
  title = {Is Your Code Generated by {ChatGPT} Really Correct? {R}igorous Evaluation of Large Language Models for Code Generation},
  author = {Liu, Jiawei and Xia, Chunqiu Steven and Wang, Yuyao and Zhang, Lingming},
  booktitle = {Advances in Neural Information Processing Systems},
  volume = {36},
  pages = {21558--21572},
  year = {2023},
  publisher = {Curran Associates, Inc.},
  doi = {10.52202/075280-0943},
  url = {https://proceedings.neurips.cc/paper_files/paper/2023/file/43e9d647ccd3e4b7b5baab53f0368686-Paper-Conference.pdf}
}

@inproceedings{lu2022scienceqa,
  title = {Learn to Explain: Multimodal Reasoning via Thought Chains for Science Question Answering},
  author = {Lu, Pan and Mishra, Swaroop and Xia, Tanglin and Qiu, Liang and Chang, Kai-Wei and Zhu, Song-Chun and Tafjord, Oyvind and Clark, Peter and Kalyan, Ashwin},
  booktitle = {Advances in Neural Information Processing Systems},
  volume = {35},
  pages = {2507--2521},
  year = {2022},
  publisher = {Curran Associates, Inc.},
  doi = {10.52202/068431-0182},
  url = {https://proceedings.neurips.cc/paper_files/paper/2022/file/11332b6b6cf4485b84afadb1352d3a9a-Paper-Conference.pdf}
}

@inproceedings{khandekar2024medcalcbench,
  title = {{MedCalc-Bench}: Evaluating Large Language Models for Medical Calculations},
  author = {Khandekar, Nikhil and Jin, Qiao and Xiong, Guangzhi and Dunn, Soren and Applebaum, Serina S and Anwar, Zain and Sarfo-Gyamfi, Maame and Safranek, Conrad W and Anwar, Abid A and Zhang, Andrew and Gilson, Aidan and Singer, Maxwell B and Dave, Amisha and Taylor, Andrew and Zhang, Aidong and Chen, Qingyu and Lu, Zhiyong},
  booktitle = {Advances in Neural Information Processing Systems},
  volume = {37},
  pages = {84730--84745},
  year = {2024},
  publisher = {Curran Associates, Inc.},
  doi = {10.52202/079017-2690},
  url = {https://proceedings.neurips.cc/paper_files/paper/2024/file/99e81750f3fdfcaf9613db2dbf4bd623-Paper-Datasets_and_Benchmarks_Track.pdf}
}
\clearpage
\appendix
\label{app:start}
\section{Six-Model Result Records and Training Protocol}
\label{app:six_model_records}
The six-model score record covers twelve arms and eleven evaluation configurations. Results summarize seeds 20260729, 20260730, and 20260731 and include Overall alongside the eleven task scores. The record contains 864 absolute values and 864 base-relative changes, including 792 task means and 72 Overall means. For each model, the ten single-category training configurations are separate from the \texttt{balanced\_mix\_10} reference and the base checkpoint.

The score and change records retain the precision exposed by the result plots. Values were transcribed from those plots and checked independently using OCR and arithmetic. Absolute scores minus the corresponding base and the independently printed changes agree within 0.1 pp; Overall also agrees with the mean of the eleven task scores within 0.1 pp. Such differences arise from separate rounding. Zero changes, including negative zero, are retained in the CSV and do not establish exact equality.

The accompanying CSVs separate these reported values from derived summaries. \texttt{absolute\_scores.csv} and \texttt{delta\_vs\_base.csv} each contain 72 model--arm rows. \texttt{all\_results\_long.csv} contains 864 model--arm--metric rows with both values, seed identifiers, and source-plot filenames. Family means in Table~\ref{tab:six_model_results} are derived from the reported task means; task-response figures use the reported deltas. These operations do not recover hidden decimal precision or per-seed variance.

\subsection{Sampling and training configuration}
Sampling balances example counts within the shared primary-label partition. Each single-category training configuration contains 1,706 examples, the size of the smallest category, \catlabel{\catGS}{greedy search and optimization}, sampled without replacement from the 35,974-example cleaned KodCode pool. The balanced arm draws from these single-category subsets: six categories contribute 171 examples each and four contribute 170, again without replacement, for $6\times171+4\times170=1{,}706$ examples. Auxiliary labels are not used to reselect examples. One epoch gives 107 optimizer steps under the recorded batch settings; input and supervised-token totals are not matched.

Training uses native chat templates, and evaluation uses the final adapter. Table~\ref{tab:six_training_configuration} records the sampling, optimization, and implementation settings. Four GPUs schedule independent jobs rather than one four-GPU data-parallel run.

\begin{table}[!htbp]
\centering\small
\caption{\textbf{Documented sampling and training configuration.} Every trained arm has the same example count and update count; token totals can differ. Base receives no additional training.}
\label{tab:six_training_configuration}
\setlength{\tabcolsep}{4pt}\renewcommand{\arraystretch}{1.08}
\begin{tabularx}{\textwidth}{@{}p{0.24\textwidth}Y@{}}
\toprule
Setting & Recorded value\\\midrule
Source and labels & 35,974 cleaned KodCode training examples; selection by primary category only.\\
Single-category configurations & 1,706 examples each; sampling without replacement.\\
Balanced mixture & 1,706 examples drawn from the single-category subsets: six quotas of 171 and four of 170.\\
Sampling seed & 20260730; no auxiliary-label reselection.\\
Input and loss & Native model chat template; assistant-only loss.\\
LoRA & Rank 16, scaling parameter 32, dropout 0.05; attention query, key, value, and output projections.\\
Learning rate & $2\times10^{-5}$.\\
Epochs and steps & One epoch; 107 optimizer updates per trained arm.\\
Batch size & One example per GPU; gradient accumulation 16; effective batch size 16.\\
Schedule & Cosine; warmup ratio 0.03; weight decay 0.01.\\
Precision and length & BF16; maximum sequence length 4,096.\\
Checkpoint choice & Final adapter; no selection by test score.\\
Hardware & Four H100 80GB GPUs running independent single-GPU jobs, not four-GPU DDP.\\
Software & PyTorch 2.7.1, Transformers 5.10.2, PEFT 0.19.1; AOSS-restored environment.\\
\bottomrule
\end{tabularx}
\end{table}
\FloatBarrier

\subsection{Evaluation configuration and record coverage}
Evaluation uses task-specific prompts, answer extraction, and scoring implementations. Code generation uses pass@1. ARC-Challenge, LegalBench, and ScienceQA use option log-probability scoring; the other configurations use generation. Table~\ref{tab:six_evaluation_configuration} lists the effective scoring counts and generation limits. The main analysis aggregates saved scores without regenerating answers. Appendix~\ref{app:imp_qa_audit} checks the saved ARC-Challenge and ScienceQA predictions against their option scores and gold answers. Appendix~\ref{app:finding1_audit} extends the single-seed audit to QA composition, mathematical answer extraction, and code execution records. Both retain the original scores; the alternative GSM8K extraction rule is reported separately as a sensitivity analysis.

\begin{table}[!htbp]
\centering\small
\caption{\textbf{Documented evaluation configuration.} Counts are the effective scored problems for each configuration. A dash indicates choice scoring, with no generation limit applicable.}
\label{tab:six_evaluation_configuration}
\setlength{\tabcolsep}{4pt}\renewcommand{\arraystretch}{1.08}
\begin{tabularx}{\textwidth}{@{}lrlYr@{}}
\toprule
Configuration & Scored $n$ & Metric & Scoring mode & Max. tokens\\\midrule
ARC-Challenge & 1,172 & Accuracy & Option log probability & ---\\
FinQA & 1,133 & Accuracy & Generated answer & 128\\
HumanEval & 164 & Pass@1 & Generated code & 2,048\\
LegalBench & 1,689 & Accuracy & Option log probability & ---\\
MATH-M & 105 & Accuracy & Generated answer & 512\\
MBPP+ & 378 & Pass@1 & Generated code & 2,048\\
MedCalc & 1,100 & Accuracy & Generated answer & 128\\
MATH-H & 262 & Accuracy & Generated answer & 512\\
GSM8K & 1,319 & Accuracy & Generated answer & 256\\
MBPP-Simple & 257 & Pass@1 & Generated code & 2,048\\
ScienceQA & 2,224 & Accuracy & Option log probability & ---\\
\bottomrule
\end{tabularx}
\end{table}

Generation and choice scoring use separate batch sizes. Decoding settings are fixed: generation uses temperature 0.2, top-$p$ 0.95, and batch size four; choice scoring uses batch size 24. The Llama-3.2 template date is fixed to 22 Sep 2026. The environment does not install the optional \texttt{math-verify} package, so mathematical answers use the evaluator's fallback matching logic and retain its parser compatibility limits.

Overall weights the eleven task percentages equally while retaining their distinct scoring populations. The source field \texttt{overall\_mean\_no\_health} is the arithmetic mean of the eleven task columns; HealthBench, APPS, and PlanBench are excluded. FinQA starts with 1,147 records and skips fourteen without target answers, leaving 1,133 scored records. MBPP-Simple and MBPP+ may overlap, so their counts cannot be summed as a disjoint MBPP test set. Upstream score CSVs are documented at ten decimal places, while the result plots expose the one-decimal values retained in this paper's extracted CSVs.

The additional audit record supplies unrounded scores, saved QA predictions, training-file hashes, and token-budget records for seed 20260730 (Appendix~\ref{app:imp_qa_audit}). Exact replication of the reported three-seed experiment still requires corresponding records for seeds 20260729 and 20260731, including confirmation of sampling and hyperparameters, mixture quotas, optimizer implementation, checkpoint/tokenizer/dataset revisions, token exposure, and unrounded scores. The single-seed audit does not fill these gaps or estimate seed variability.
\FloatBarrier

\clearpage
\section{Aggregate and Complete Task-Response Maps}
\label{app:six_model_scores}
\subsection{Task-family scores for all training recipes}
\label{app:rq1_family_scores}
Table~\ref{tab:six_model_results} retains the complete task-family comparison for Base, the balanced mixture, and all ten single-category recipes. Figures~\ref{fig:rq1_qa_tasks} and~\ref{fig:rq1_code_math} in Section~\ref{sec:rq1} focus on Base versus the balanced mixture and separate the code and mathematics means.

\begin{table}[H]
\centering
\caption{\textbf{Code post-training transfers across tasks, with uneven gains across models.} Entries are three-seed mean task-group scores (\%), averaged from the reported one-decimal task scores. All 60 single-category training configurations improve the QA mean (a), whereas only 24 improve the combined code-and-mathematics mean (b). Blue/red shading indicates \sixlegend{4F8FBC}{higher}/\sixlegend{E8B0B0}{lower} performance relative to the same backbone's Base, using averages of the reported base-relative deltas. Panel (a) saturates at $+3.5$ pp and panel (b) at $\pm3$ pp. Base rows are unshaded; the balanced mix matches each single-category configuration's example budget.}
\label{tab:six_model_results}
\vspace{6pt}
\begingroup
\fontsize{7.2}{8.5}\selectfont
\setlength{\tabcolsep}{0pt}
\renewcommand{\arraystretch}{1}
\arrayrulecolor{tabrule}
\newcommand{\sixrow}{\rule[-3.3pt]{0pt}{11.5pt}}
\newcommand{\sixhead}[2]{\begin{tabular}[c]{@{}c@{}}#1\\[-1.3pt]{\fontsize{6.8}{7.3}\selectfont\bfseries #2}\end{tabular}}
\begin{tabular}{@{}p{117pt}*{6}{>{\centering\arraybackslash}p{\dimexpr(\textwidth-117pt)/6\relax}}@{}}
\toprule
\multicolumn{7}{@{}l@{}}{\rule[-3pt]{0pt}{13pt}\textbf{(a) QA (5 configurations)}}\\
\rowcolor{white} \textbf{Training recipe} & \sixhead{\includegraphics[height=8pt]{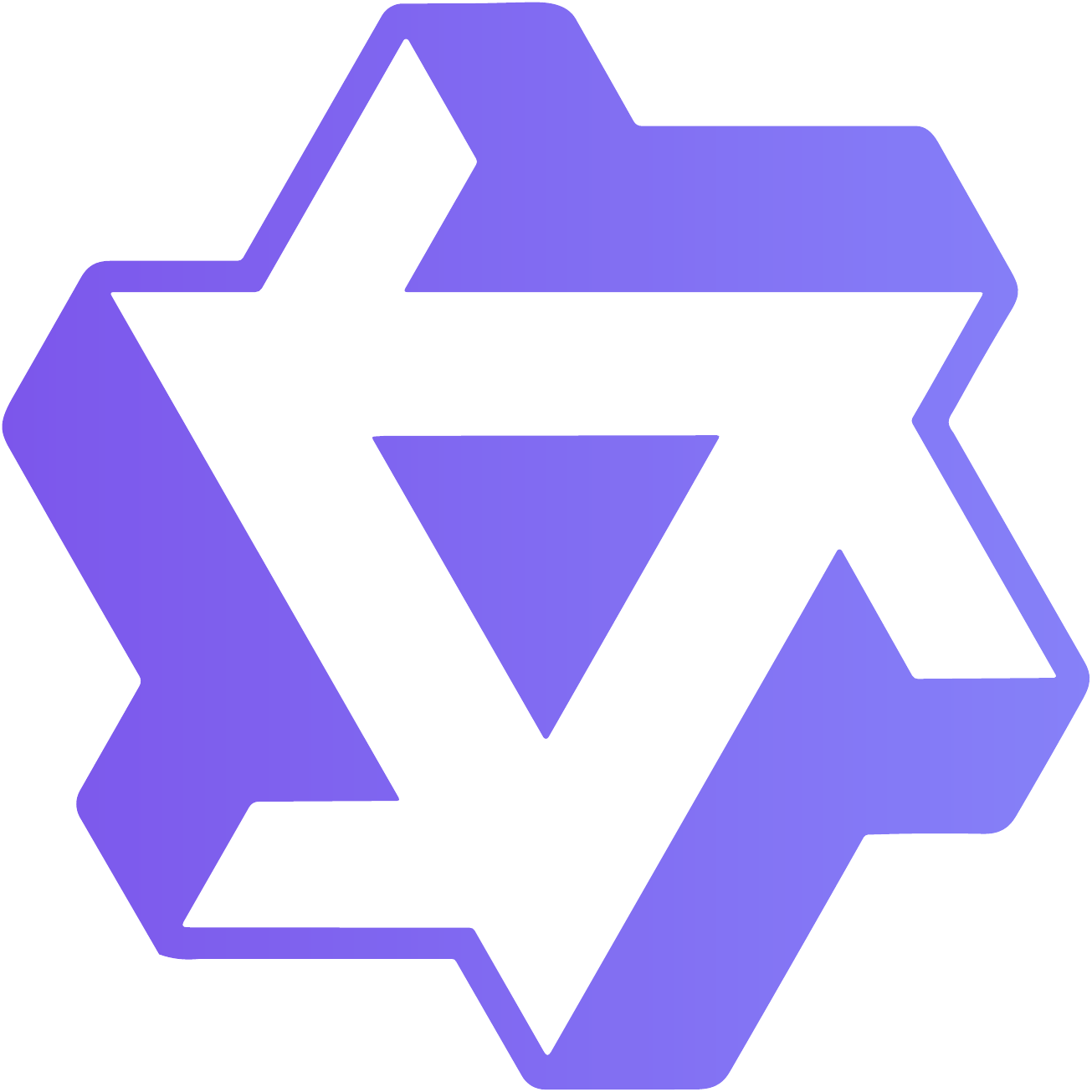}}{Qwen2.5-7B} & \sixhead{\includegraphics[height=8pt]{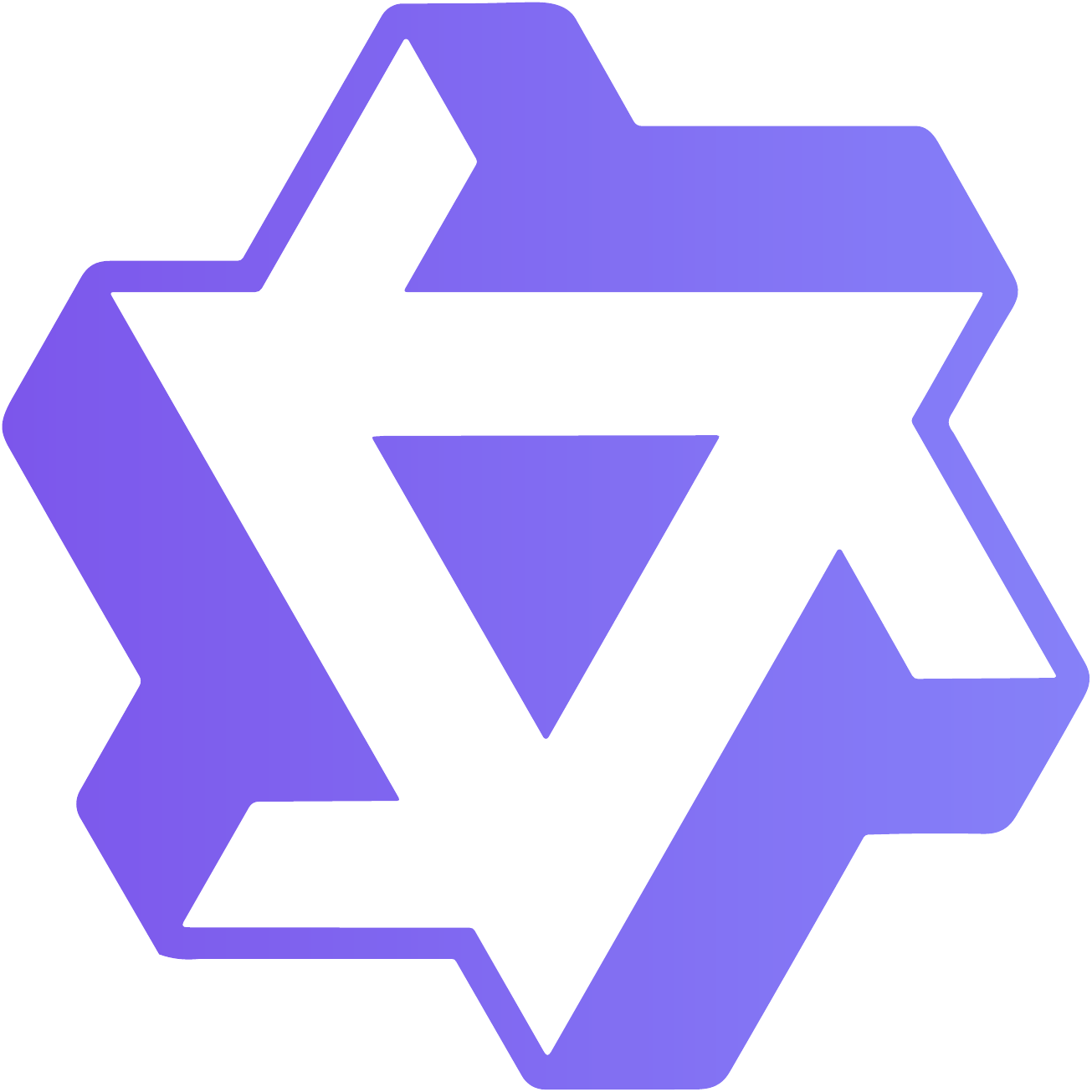}}{Qwen3-4B} & \sixhead{\includegraphics[height=8pt]{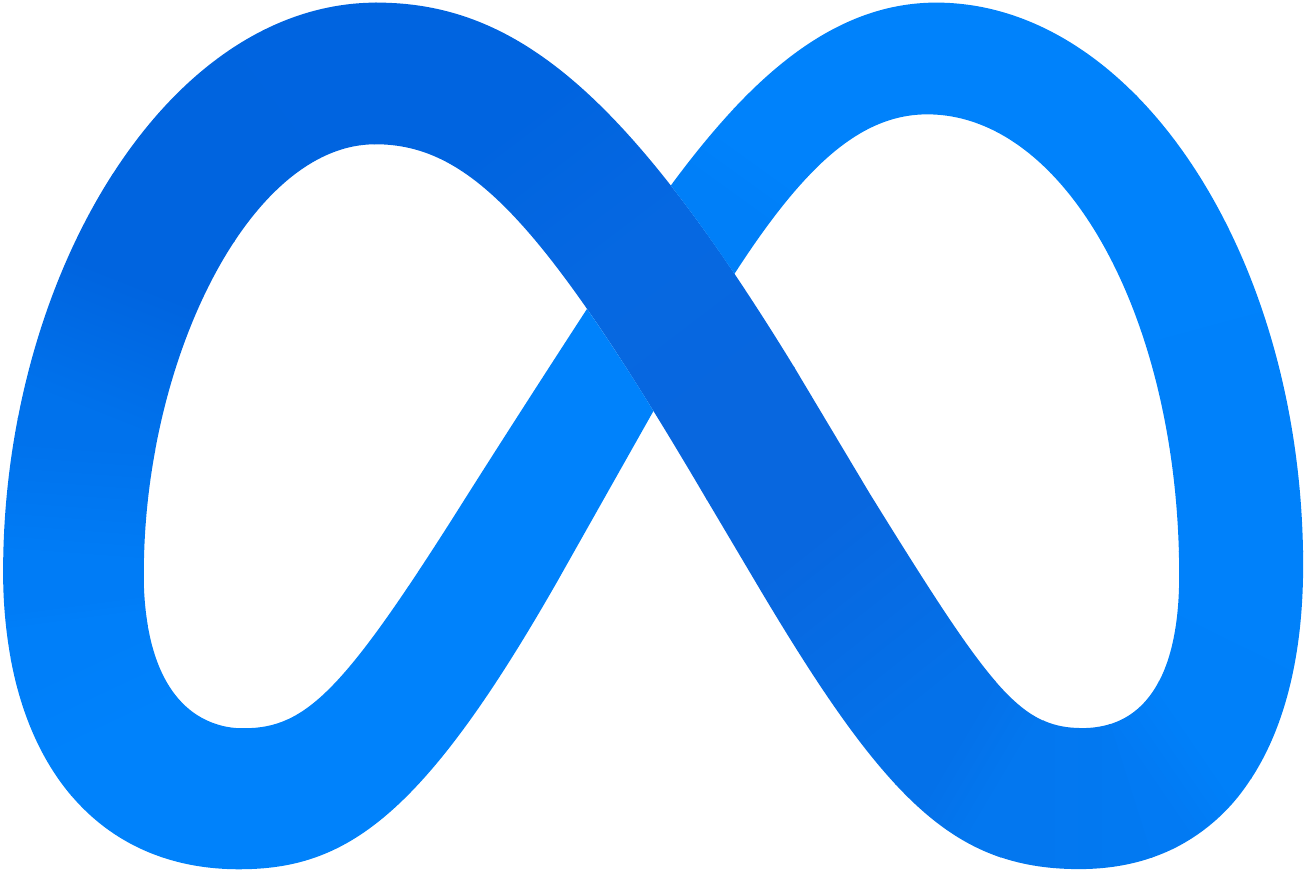}}{Llama3.1-8B} & \sixhead{\includegraphics[height=8pt]{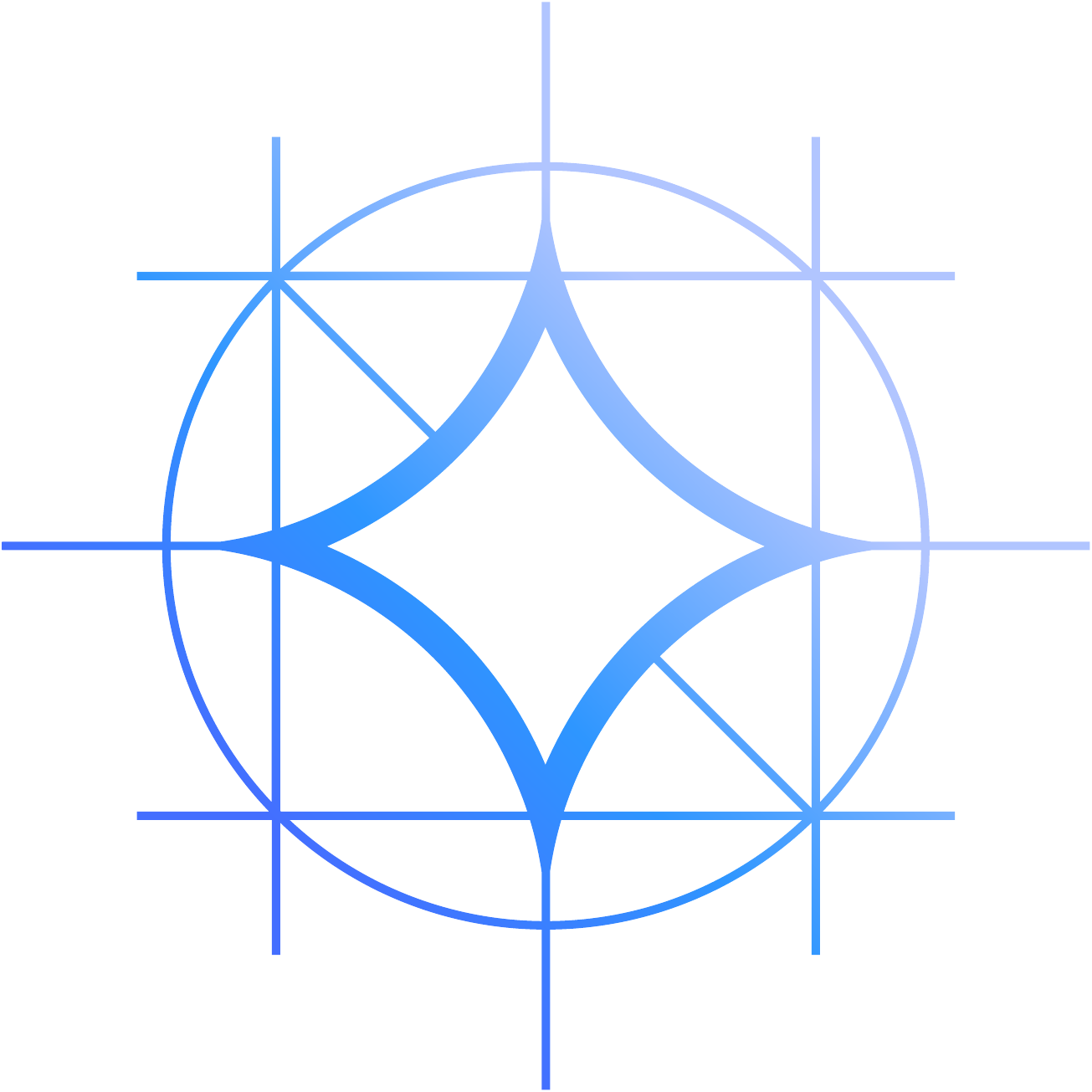}}{Gemma2-9B} & \sixhead{\includegraphics[height=8pt]{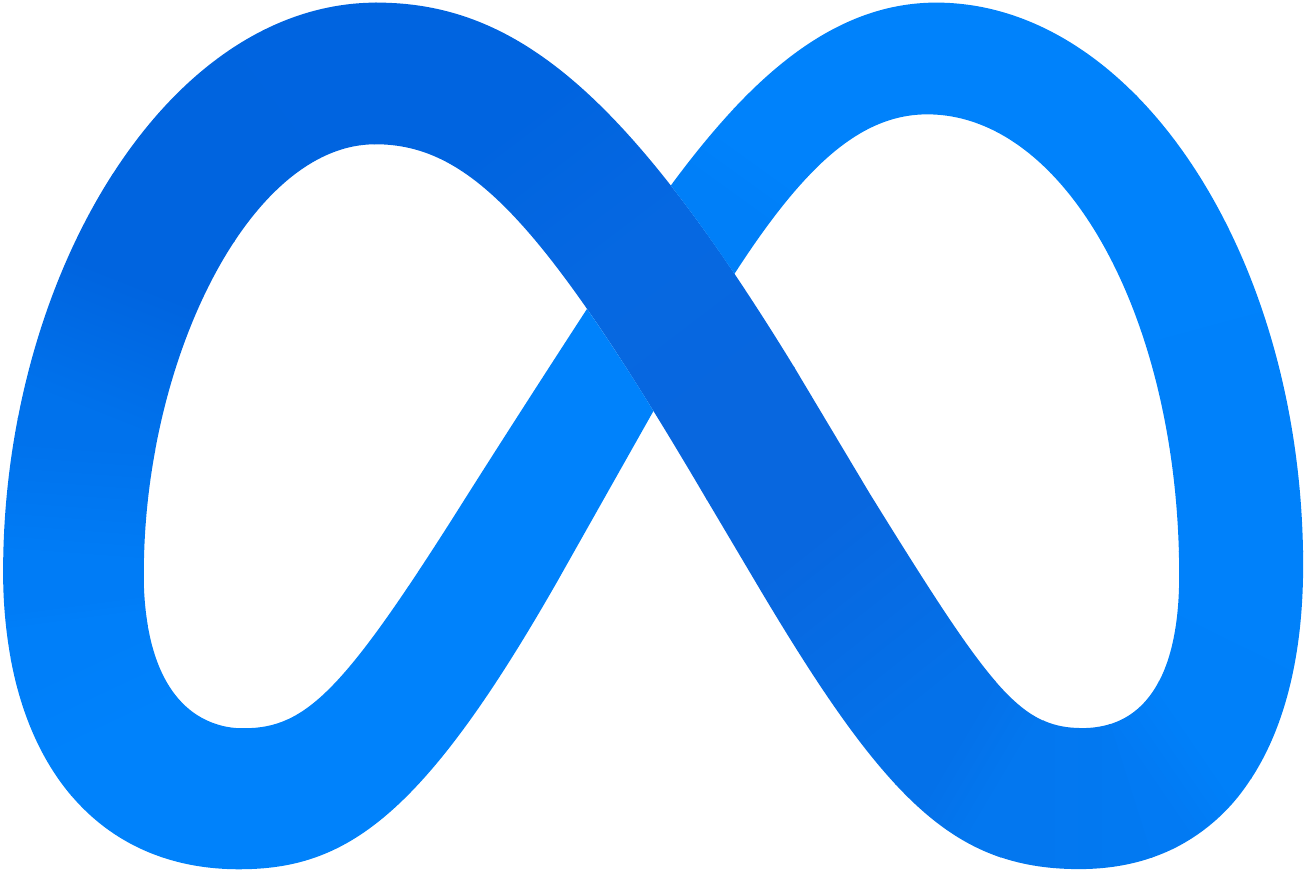}}{Llama3.2-3B} & \sixhead{\includegraphics[height=8pt]{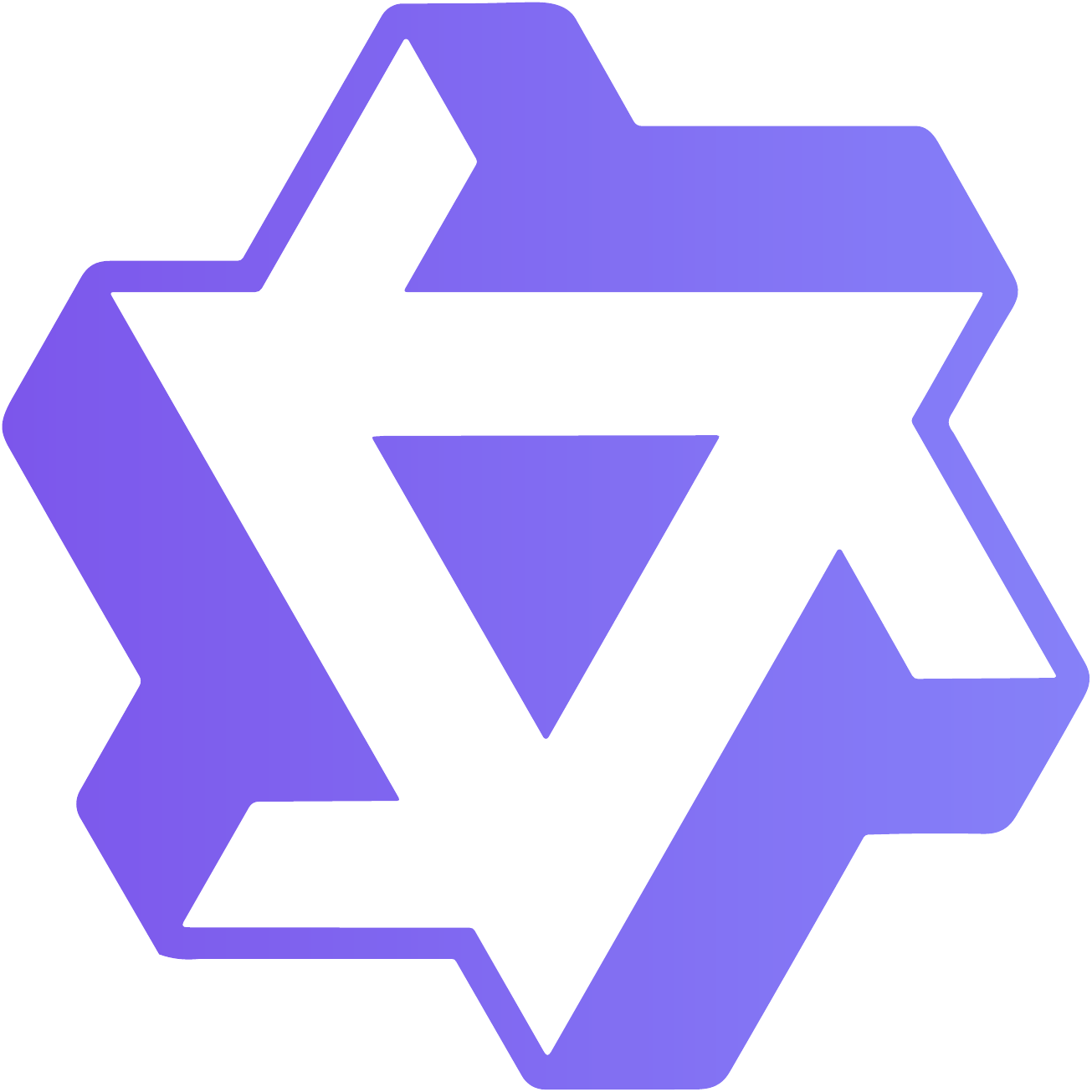}}{Qwen1.5-7B} \\
\midrule
\sixrow \textbf{Base} & 43.3 & 43.1 & 44.8 & 44.5 & 36.2 & 36.2 \\
\midrule
\sixrow \textbf{Balanced mix} & \cellcolor[HTML]{6DA2C7}46.2 & \cellcolor[HTML]{8BB5D3}45.4 & \cellcolor[HTML]{E3EDF4}45.4 & \cellcolor[HTML]{8DB7D4}46.7 & \cellcolor[HTML]{B6D0E3}37.7 & \cellcolor[HTML]{D7E5F0}36.9 \\
\sixrow \catMath\enspace Math / NT & \cellcolor[HTML]{6DA2C7}46.2 & \cellcolor[HTML]{71A5C9}45.9 & \cellcolor[HTML]{ECF3F8}45.2 & \cellcolor[HTML]{7EADCE}47.0 & \cellcolor[HTML]{D2E2EE}37.2 & \cellcolor[HTML]{DAE7F1}36.9 \\
\sixrow \catDP\enspace Dynamic programming & \cellcolor[HTML]{76A8CB}46.0 & \cellcolor[HTML]{8CB6D3}45.4 & \cellcolor[HTML]{E2ECF4}45.4 & \cellcolor[HTML]{8EB7D4}46.7 & \cellcolor[HTML]{B9D2E4}37.6 & \cellcolor[HTML]{F0F5F9}36.4 \\
\sixrow \catGR\enspace Graph / stateful & \cellcolor[HTML]{79AACC}45.9 & \cellcolor[HTML]{A5C6DD}44.9 & \cellcolor[HTML]{E6EFF5}45.3 & \cellcolor[HTML]{88B3D2}46.8 & \cellcolor[HTML]{C3D9E8}37.5 & \cellcolor[HTML]{EFF5F9}36.5 \\
\sixrow \catGS\enspace Greedy / search & \cellcolor[HTML]{6EA3C8}46.2 & \cellcolor[HTML]{8AB5D3}45.5 & \cellcolor[HTML]{E4EEF5}45.4 & \cellcolor[HTML]{91B9D5}46.6 & \cellcolor[HTML]{BBD3E5}37.6 & \cellcolor[HTML]{E8F0F6}36.6 \\
\sixrow \catIS\enspace Impl. / utilities & \cellcolor[HTML]{4F8FBC}46.8 & \cellcolor[HTML]{6EA3C8}46.0 & \cellcolor[HTML]{D0E1ED}45.8 & \cellcolor[HTML]{81AFCF}47.0 & \cellcolor[HTML]{B1CDE1}37.8 & \cellcolor[HTML]{E1ECF4}36.7 \\
\sixrow \catHash\enspace Hashing / sets & \cellcolor[HTML]{80AECF}45.8 & \cellcolor[HTML]{7CACCD}45.7 & \cellcolor[HTML]{E8F0F6}45.3 & \cellcolor[HTML]{90B9D5}46.7 & \cellcolor[HTML]{BCD4E5}37.6 & \cellcolor[HTML]{F5F9FB}36.3 \\
\sixrow \catRange\enspace Range / matrix & \cellcolor[HTML]{7DACCE}45.9 & \cellcolor[HTML]{94BBD6}45.2 & \cellcolor[HTML]{F0F5F9}45.1 & \cellcolor[HTML]{92BAD6}46.6 & \cellcolor[HTML]{B5D0E3}37.8 & \cellcolor[HTML]{E6EFF5}36.6 \\
\sixrow \catSeq\enspace Sequence transform. & \cellcolor[HTML]{75A7CB}46.1 & \cellcolor[HTML]{84B1D0}45.5 & \cellcolor[HTML]{D4E3EF}45.7 & \cellcolor[HTML]{91B9D5}46.6 & \cellcolor[HTML]{A4C5DD}38.1 & \cellcolor[HTML]{E0EBF3}36.7 \\
\sixrow \catSort\enspace Sorting / ordered & \cellcolor[HTML]{7BABCD}45.9 & \cellcolor[HTML]{94BBD6}45.2 & \cellcolor[HTML]{EBF2F7}45.2 & \cellcolor[HTML]{94BBD6}46.6 & \cellcolor[HTML]{ACCADF}37.9 & \cellcolor[HTML]{E4EEF5}36.7 \\
\sixrow \catStr\enspace String / parsing & \cellcolor[HTML]{7AABCC}45.9 & \cellcolor[HTML]{81AFCF}45.6 & \cellcolor[HTML]{E7F0F6}45.3 & \cellcolor[HTML]{96BCD7}46.6 & \cellcolor[HTML]{A8C7DE}38.0 & \cellcolor[HTML]{E9F1F7}36.6 \\
\midrule
\addlinespace[2pt]
\multicolumn{7}{@{}l@{}}{\rule[-3pt]{0pt}{13pt}\textbf{(b) Code \& math (6 configurations)}}\\
\rowcolor{white} \textbf{Training recipe} & \sixhead{\includegraphics[height=8pt]{figures/model_icons/Qwen2.5-7B.pdf}}{Qwen2.5-7B} & \sixhead{\includegraphics[height=8pt]{figures/model_icons/Qwen3-4B.pdf}}{Qwen3-4B} & \sixhead{\includegraphics[height=8pt]{figures/model_icons/Llama3.1-8B.pdf}}{Llama3.1-8B} & \sixhead{\includegraphics[height=8pt]{figures/model_icons/Gemma2-9B.pdf}}{Gemma2-9B} & \sixhead{\includegraphics[height=8pt]{figures/model_icons/Llama3.2-3B.pdf}}{Llama3.2-3B} & \sixhead{\includegraphics[height=8pt]{figures/model_icons/Qwen1.5-7B.pdf}}{Qwen1.5-7B} \\
\midrule
\sixrow \textbf{Base} & 66.2 & 65.4 & 57.1 & 60.4 & 52.1 & 38.0 \\
\midrule
\sixrow \textbf{Balanced mix} & \cellcolor[HTML]{BBD6E9}68.2 & \cellcolor[HTML]{ECBEBE}62.9 & \cellcolor[HTML]{FEFDFD}57.0 & \cellcolor[HTML]{F5DEDE}59.1 & \cellcolor[HTML]{F4DADA}50.7 & \cellcolor[HTML]{EAF2F8}38.6 \\
\sixrow \catMath\enspace Math / NT & \cellcolor[HTML]{CCE0EE}67.7 & \cellcolor[HTML]{F5DEDE}64.1 & \cellcolor[HTML]{F8E6E6}56.1 & \cellcolor[HTML]{F9EAEA}59.6 & \cellcolor[HTML]{F7E4E4}51.1 & \cellcolor[HTML]{BAD5E8}40.1 \\
\sixrow \catDP\enspace Dynamic programming & \cellcolor[HTML]{DDEAF4}67.3 & \cellcolor[HTML]{F8E7E7}64.4 & \cellcolor[HTML]{F6E0E0}55.9 & \cellcolor[HTML]{F0CBCB}58.4 & \cellcolor[HTML]{FDFAFA}51.9 & \cellcolor[HTML]{EEF5FA}38.5 \\
\sixrow \catGR\enspace Graph / stateful & \cellcolor[HTML]{C8DDED}67.9 & \cellcolor[HTML]{F7E4E4}64.3 & \cellcolor[HTML]{F6F9FC}57.3 & \cellcolor[HTML]{F4D8D8}58.9 & \cellcolor[HTML]{F4DBDB}50.7 & \cellcolor[HTML]{CFE2EF}39.4 \\
\sixrow \catGS\enspace Greedy / search & \cellcolor[HTML]{D7E7F2}67.4 & \cellcolor[HTML]{ECBDBD}62.9 & \cellcolor[HTML]{F6DFDF}55.9 & \cellcolor[HTML]{F9EBEB}59.6 & \cellcolor[HTML]{F5DEDE}50.8 & \cellcolor[HTML]{F7FAFC}38.2 \\
\sixrow \catIS\enspace Impl. / utilities & \cellcolor[HTML]{ADCDE4}68.7 & \cellcolor[HTML]{F1D0D0}63.6 & \cellcolor[HTML]{EDF4F9}57.6 & \cellcolor[HTML]{FEFDFD}60.4 & \cellcolor[HTML]{EDC3C3}49.8 & \cellcolor[HTML]{F2F7FB}38.4 \\
\sixrow \catHash\enspace Hashing / sets & \cellcolor[HTML]{B5D2E7}68.4 & \cellcolor[HTML]{F9EAEA}64.6 & \cellcolor[HTML]{FCF4F4}56.7 & \cellcolor[HTML]{F6E1E1}59.3 & \cellcolor[HTML]{F7E5E5}51.1 & \cellcolor[HTML]{C5DCEC}39.7 \\
\sixrow \catRange\enspace Range / matrix & \cellcolor[HTML]{BFD8EA}68.1 & \cellcolor[HTML]{ECBEBE}62.9 & \cellcolor[HTML]{FDFAFA}56.9 & \cellcolor[HTML]{F3D5D5}58.8 & \cellcolor[HTML]{F7E5E5}51.1 & \cellcolor[HTML]{FCF3F3}37.5 \\
\sixrow \catSeq\enspace Sequence transform. & \cellcolor[HTML]{C7DDEC}67.9 & \cellcolor[HTML]{FCF4F4}65.0 & \cellcolor[HTML]{FDF7F7}56.8 & \cellcolor[HTML]{F7E4E4}59.4 & \cellcolor[HTML]{EBBABA}49.5 & \cellcolor[HTML]{F7FAFC}38.2 \\
\sixrow \catSort\enspace Sorting / ordered & \cellcolor[HTML]{D4E5F1}67.5 & \cellcolor[HTML]{F7E5E5}64.3 & \cellcolor[HTML]{DDEAF4}58.1 & \cellcolor[HTML]{F3D6D6}58.9 & \cellcolor[HTML]{F5F9FC}52.4 & \cellcolor[HTML]{F6FAFC}38.3 \\
\sixrow \catStr\enspace String / parsing & \cellcolor[HTML]{ACCCE4}68.7 & \cellcolor[HTML]{EBBBBB}62.8 & \cellcolor[HTML]{F8E7E7}56.2 & \cellcolor[HTML]{F7FAFC}60.6 & \cellcolor[HTML]{FAEDED}51.4 & \cellcolor[HTML]{F6F9FC}38.3 \\
\bottomrule
\end{tabular}
\endgroup
\end{table}
\FloatBarrier

\subsection{Aggregate responses across backbones}
\label{app:aggregate_results}
\begin{figure}[!htbp]
\centering
\includegraphics[width=\textwidth]{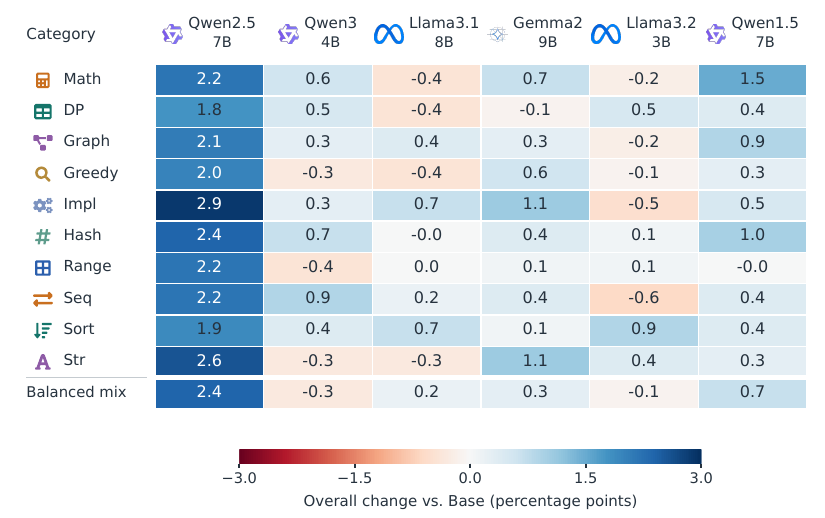}
\caption{\textbf{Overall responses depend on both the backbone and the training category.} Cells show the reported three-seed mean Overall change from the corresponding base checkpoint, in percentage points, for ten single-category training configurations and the balanced ten-category mixture. Blue denotes improvement and red denotes decline, with one shared scale. Category icons follow Table~\ref{tab:taxonomy}. Values are reported to one decimal place.}
\label{fig:six_overall}
\end{figure}

Single-category training improves Overall in 44 of the 60 model--category combinations. Figure~\ref{fig:six_overall} reports changes from each model's base checkpoint: thirteen combinations decrease and three round to zero. The distribution differs substantially across backbones. All ten categories improve Qwen2.5-7B, with gains of $+1.8$ to $+2.9$ pp. Nine improve Gemma-2-9B and Qwen1.5-7B, seven improve Qwen3-4B, five improve Llama-3.2-3B, and four improve Llama-3.1-8B. These counts describe the directions of the reported means.

The magnitude of these changes also differs across backbones: the strongest aggregate gains occur on Qwen2.5, while the Llama models show smaller changes of both signs. \catlabel{\catIS}{Direct implementation} raises Qwen2.5 Overall from $55.8\%$ to $58.7\%$. By comparison, category responses span $-0.4$ to $+0.7$ pp on Llama-3.1 and $-0.6$ to $+0.9$ pp on Llama-3.2. An aggregate close to base can nevertheless combine sizable opposing task changes. Table~\ref{tab:six_model_results} separates QA from code and mathematics, and Figure~\ref{fig:six_task_map} resolves each task. These views identify which task responses contribute to the aggregate change.

\FloatBarrier
\subsection{Complete per-backbone task responses}
To expose the task changes underlying these aggregates, the per-backbone maps retain all category--task responses and the balanced-mixture reference. Every cell is a reported three-seed change from that backbone's base checkpoint in percentage points, shown at the source's one-decimal precision. All maps use the same zero-centered scale, with blue denoting improvement and red denoting decline. Base-relative deltas are plotted directly rather than reconstructed from rounded absolute scores. Category icons match Table~\ref{tab:taxonomy}; the final row is the balanced ten-category mixture. The companion CSVs retain all absolute scores as well as these changes.

\begin{figure}[p]
\centering
\includegraphics[width=0.92\textwidth]{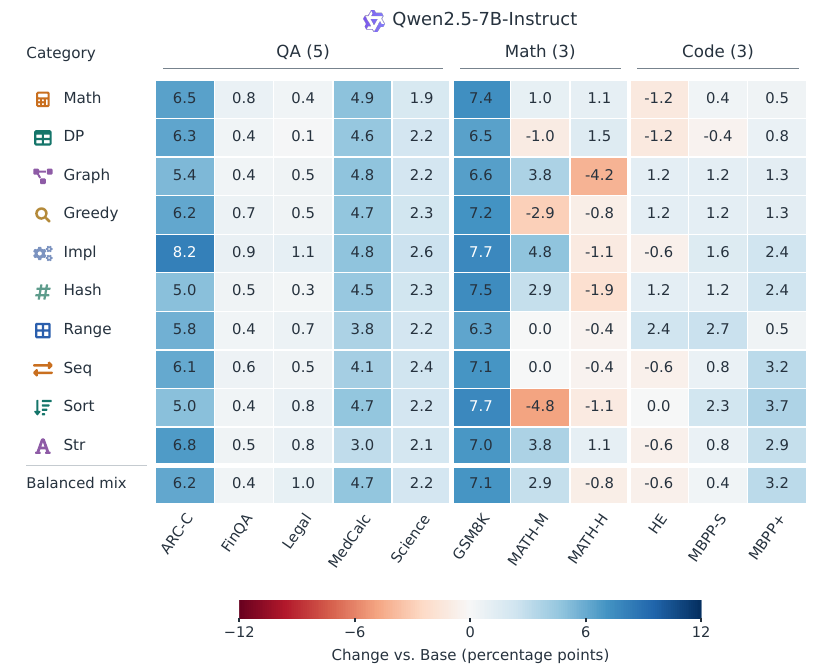}
\caption{\textbf{Qwen2.5-7B-Instruct: complete task responses.}}
\label{fig:six_qwen25_full}
\end{figure}

\begin{figure}[p]
\centering
\includegraphics[width=0.92\textwidth]{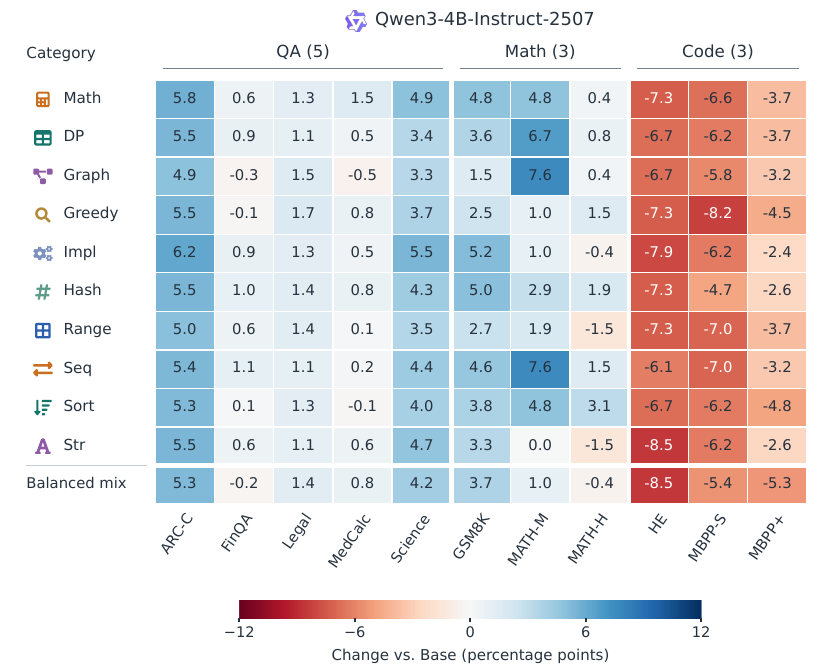}
\caption{\textbf{Qwen3-4B-Instruct-2507: complete task responses.}}
\label{fig:six_qwen3_full}
\end{figure}

\begin{figure}[p]
\centering
\includegraphics[width=0.92\textwidth]{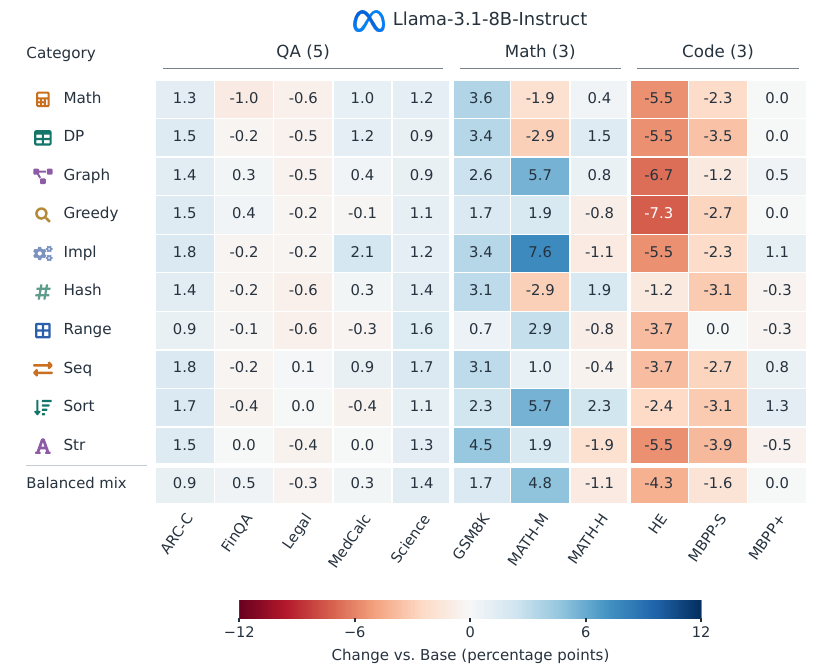}
\caption{\textbf{Llama-3.1-8B-Instruct: complete task responses.}}
\label{fig:six_llama31_full}
\end{figure}

\begin{figure}[p]
\centering
\includegraphics[width=0.92\textwidth]{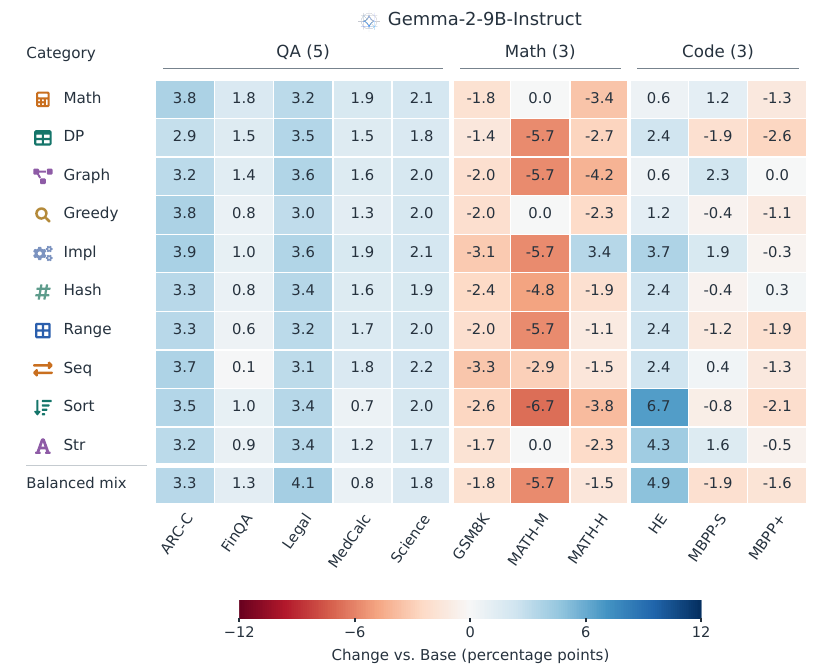}
\caption{\textbf{Gemma-2-9B-Instruct: complete task responses.}}
\label{fig:six_gemma2_full}
\end{figure}

\begin{figure}[p]
\centering
\includegraphics[width=0.92\textwidth]{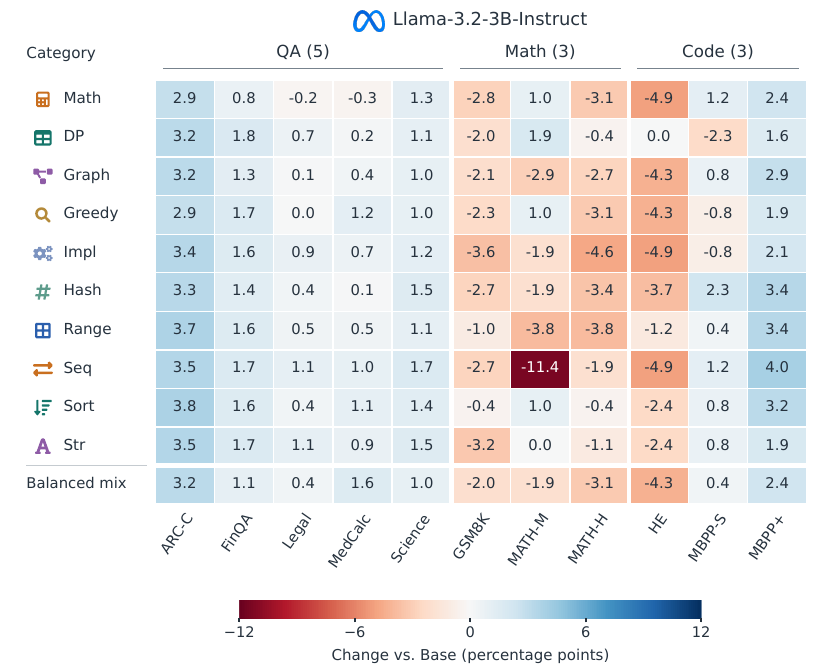}
\caption{\textbf{Llama-3.2-3B-Instruct: complete task responses.}}
\label{fig:six_llama32_full}
\end{figure}

\begin{figure}[p]
\centering
\includegraphics[width=0.92\textwidth]{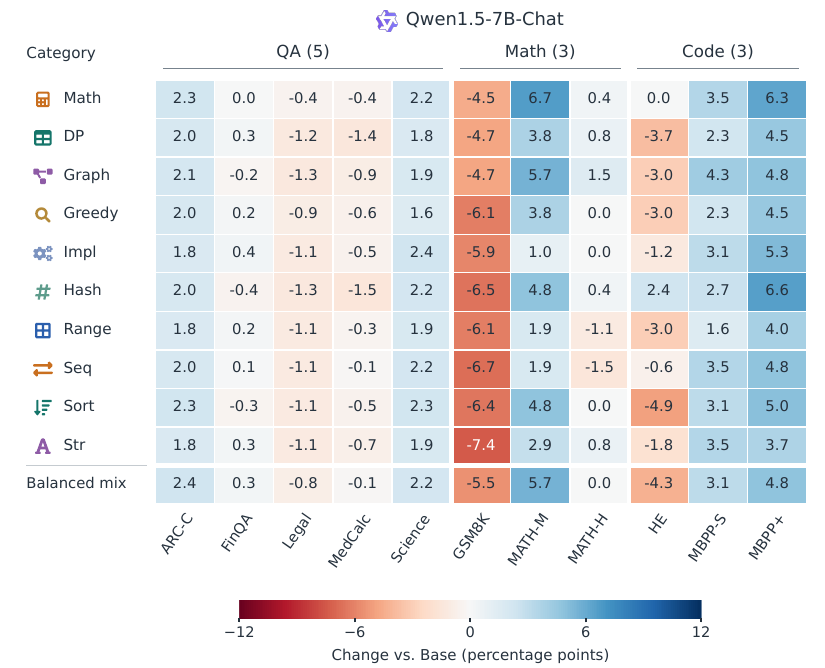}
\caption{\textbf{Qwen1.5-7B-Chat: complete task responses.}}
\label{fig:six_qwen15_full}
\end{figure}
\FloatBarrier
\clearpage
\section{Compact-Mixture Experiments and Further Validation}
\label{app:selection_protocol}
This experiment tests jointly trained mixtures suggested by the category responses. The initial experiment compares four compact recipes with a newly trained full-corpus reference on all six backbones. The comparisons are exploratory: model-specific categories were chosen using earlier results on the same eleven test configurations. The shared three-category recipe, shown in Figure~\ref{fig:compact_mixture}b, was chosen heuristically from favorable single-category responses and complementary code patterns: \catlabel{\catIS}{direct implementation (Imp)}, \catlabel{\catHash}{set/count manipulation (Hash)}, and \catlabel{\catSort}{ordered processing (Sort)}. \catlabel{\catIS}{Imp}'s QA-family gain is positive on all six backbones, and all three categories have high cross-backbone mean Overall scores in the preceding sweep. The recipe was fixed before mixture training; complementarity motivated testing the combination, without establishing additive effects or optimality.

\subsection{Recipes, budgets, and reporting scope}
\label{app:mixture_setup}
Table~\ref{tab:mixture_design} distinguishes the five trained arms from Base. The two Top3 recipes concatenate the earlier fixed 1,706-example category subsets and shuffle them. The balanced recipe samples from those category subsets; the proportional recipe samples within categories from the complete pool without replacement. All trained arms start from the original checkpoint; Complete is a fresh reference from this round, not an older full-training result or a continuation of a compact arm.

\begin{table}[H]
\centering\small
\caption{\textbf{Mixture training arms.} Compact arms match examples and updates, but not token exposure. Complete uses the complete 35,974-example source pool.}
\label{tab:mixture_design}
\begin{tabularx}{\textwidth}{lrrY}
\toprule
Arm & Examples & Updates & Composition \\
\midrule
Base & 0 & 0 & Original checkpoint \\
Personalized Top3 & 5,118 & 320 & Three model-specific categories, 1,706 each \\
Shared Top3 & 5,118 & 320 & \catlabel{\catIS}{Impl./util.}, \catlabel{\catHash}{Hash/sets}, \catlabel{\catSort}{Sort/order}; 1,706 each \\
Balanced & 5,118 & 320 & \catlabel{\catMath}{Math} and \catlabel{\catSeq}{Seq}: 511 each; other eight: 512 each \\
Proportional & 5,118 & 320 & Stratified sample preserving pool proportions \\
Complete & 35,974 & 2,249 & Full source pool, one epoch \\
\bottomrule
\end{tabularx}
\end{table}

Personalized Top3 takes the three highest earlier single-category Overall scores, not a QA-specific ranking. The selected sets are \catlabel{\catIS}{Imp}/\catlabel{\catStr}{Str}/\catlabel{\catHash}{Hash} (Qwen2.5), \catlabel{\catSeq}{Seq}/\catlabel{\catHash}{Hash}/\catlabel{\catMath}{Math} (Qwen3), \catlabel{\catMath}{Math}/\catlabel{\catHash}{Hash}/\catlabel{\catGR}{Graph} (Qwen1.5), \catlabel{\catSort}{Sort}/\catlabel{\catIS}{Imp}/\catlabel{\catGR}{Graph} (Llama-3.1), \catlabel{\catSort}{Sort}/\catlabel{\catDP}{DP}/\catlabel{\catStr}{Str} (Llama-3.2), and \catlabel{\catIS}{Imp}/\catlabel{\catStr}{Str}/\catlabel{\catMath}{Math} (Gemma-2). The proportional quotas are \catlabel{\catIS}{Imp} 800, \catlabel{\catHash}{Hash} 722, \catlabel{\catSeq}{Seq} 632, \catlabel{\catRange}{Range} 600, \catlabel{\catStr}{Str} 567, \catlabel{\catDP}{DP} 507, \catlabel{\catGR}{Graph} 395, \catlabel{\catSort}{Sort} 357, \catlabel{\catMath}{Math} 295, and \catlabel{\catGS}{Greedy} 243. Freezing these recipes before training does not make their evaluation independent of the earlier selection scores.

Training uses rank-16 LoRA with alpha 32, dropout 0.05, and q/k/v/o projections; learning rate $2\times10^{-5}$ with cosine decay, warmup ratio 0.03, and weight decay 0.01; effective batch size 16; one epoch; BF16; maximum length 4,096; and assistant-only loss. The final adapter is evaluated with each model's native chat template and thinking disabled. Evaluation retains the original scoring rules, temperature 0.2 and top-$p$ 0.95. Generation caps are 2,048 tokens for code, 256 for GSM8K, 512 for both MATH configurations, and 128 for FinQA/MedCalc. ARC, ScienceQA, and LegalBench use candidate likelihoods. The task denominators match Appendix~\ref{app:six_model_records}; the MATH subsets contain 105 and 262 questions.

The reported mixture means average seeds 20260730, 20260729, and 20270731, as confirmed by the authors; the third identifier intentionally differs from the earlier sweep. The accompanying report's detailed run paths and audit counts document seed 20260730. Individual scores for all seeds and raw predictions are absent from the supplied compact package, so these results do not quantify seed variability. The Overall tables preserve three decimals from the report, and task maps preserve one decimal from the source plots. The highlighted MATH contrasts use the report's finer-precision differences, not subtraction of rounded plot values. Machine-readable transcriptions, original primary score plots, and provenance notes accompany the source in \texttt{data/mixtures/}.

\clearpage
\subsection{Overall scores and matched-budget comparisons}
\label{app:mixture_overall}
\begin{table}[H]
\centering\small
\caption{\textbf{All mixture Overall means (\%).} Overall equally weights eleven task percentages, with GSM8K at 256 tokens. Bold marks the largest trained-arm mean in each row; it does not denote statistical significance or necessarily an improvement over Base.}
\label{tab:mixture_overall}
\setlength{\tabcolsep}{4pt}
\begin{tabular}{lrrrrrr}
\toprule
Model & Base & Personal. & Shared & Balanced & Proport. & Complete \\
\midrule
Qwen2.5-7B & 55.782 & 58.571 & \textbf{59.165} & 58.825 & 59.035 & 56.781 \\
Qwen3-4B & 55.242 & 56.872 & 56.361 & 57.361 & 56.651 & \textbf{57.975} \\
Qwen1.5-7B & 37.145 & 37.347 & \textbf{38.407} & 38.244 & 38.066 & 37.667 \\
Llama-3.1-8B & 51.496 & 52.077 & 51.726 & 52.363 & 51.589 & \textbf{52.900} \\
Llama-3.2-3B & 44.898 & 43.971 & \textbf{44.264} & 44.209 & 43.775 & 41.373 \\
Gemma-2-9B & 53.165 & \textbf{53.988} & 53.400 & 53.438 & 53.660 & 50.935 \\
\bottomrule
\end{tabular}
\end{table}

At least one compact recipe exceeds Complete on Overall for four of six backbones. Qwen3 and Llama-3.1 instead favor Complete; their best compact recipe is Balanced, trailing by 0.614 and 0.537 pp. Personalized exceeds Complete on three backbones, while Shared, Balanced, and Proportional each do so on four. Thus, an observed advantage over full-corpus training is not exclusive to category-based selection. The highest compact score is selected retrospectively from four candidates and is not an independently confirmed optimum.

\begin{figure}[H]
\centering
\includegraphics[width=\textwidth]{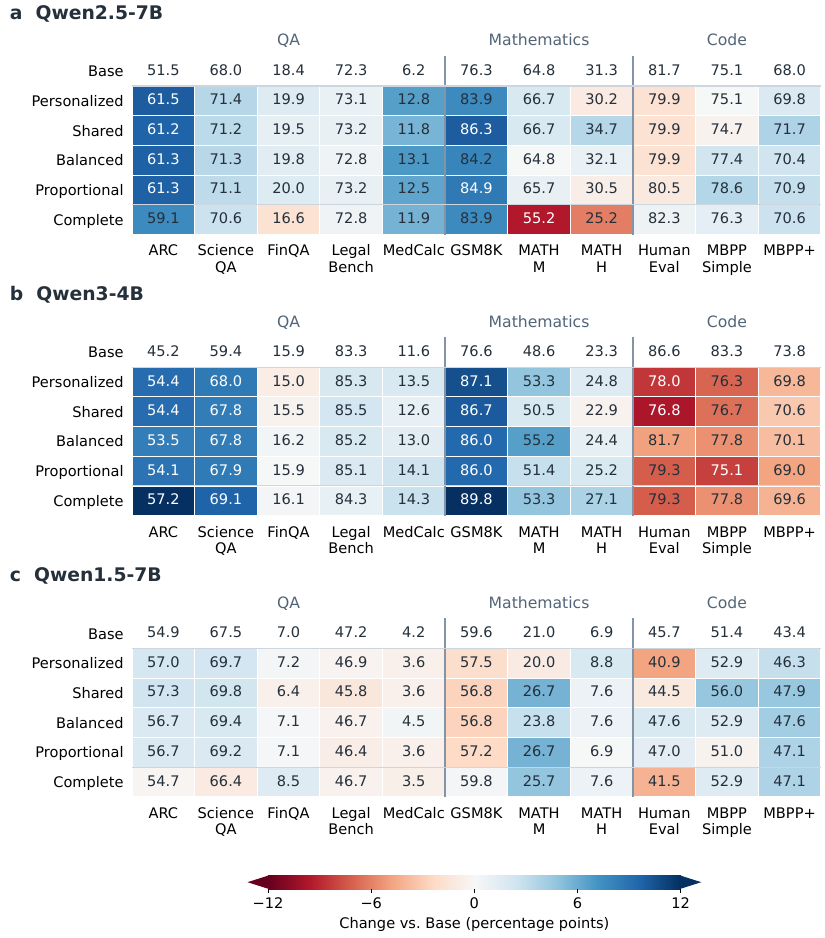}
\caption{\textbf{Complete mixture task scores: Qwen backbones.} All six original arms and eleven task configurations are shown as reported three-seed mean percentages, at the source plots' one-decimal precision. Base rows are unshaded; other cells are blue/red for gains/declines relative to the same model's Base. The shared red--blue scale matches Figure~\ref{fig:six_qwen15_full} and saturates beyond $\pm12$ pp, as indicated by the colorbar extensions. Shading uses differences between the displayed one-decimal scores; cell labels remain absolute percentages. Shared combines \catlabel{\catIS}{Impl./util.}, \catlabel{\catHash}{Hash/sets}, and \catlabel{\catSort}{Sort/order}. The primary GSM8K cap is 256 tokens; MATH M/H denote the Medium/High configurations. Overall is reported separately in Table~\ref{tab:mixture_overall}.}
\label{fig:mixture_tasks_1}
\end{figure}

\begin{figure}[H]
\centering
\includegraphics[width=\textwidth]{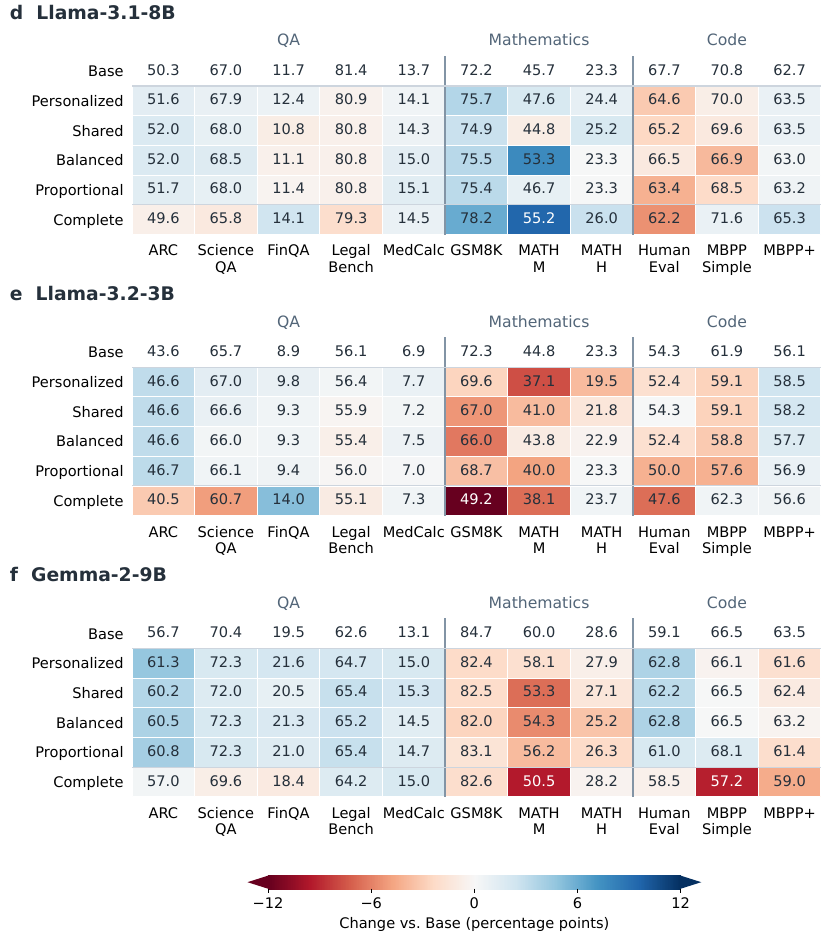}
\caption{\textbf{Complete mixture task scores: Llama and Gemma backbones.} Presentation, precision, and the $\pm12$ pp color scale follow Figure~\ref{fig:mixture_tasks_1}. Base rows remain unshaded. Base and Complete are retained alongside every compact recipe so that reduced degradation can be distinguished from improvement over the starting checkpoint. No seed-level error bars or significance tests are inferred from these means.}
\label{fig:mixture_tasks_2}
\end{figure}

Base remains essential to interpretation. All four Llama-3.2 compact arms exceed Complete yet fall below Base; Shared is 44.264 versus 44.898 for Base and 41.373 for Complete. Across all backbones, 20 of 24 compact arms raise Overall and all 24 raise the five-task QA mean. The QA-family direction does not imply gains on every constituent task. Figures~\ref{fig:mixture_tasks_1} and~\ref{fig:mixture_tasks_2} retain all task scores, including the decreases.

\begin{table}[H]
\centering\small
\caption{\textbf{Top3 recipes versus equally sized controls (Overall, pp).} Each contrast is reported at the source precision. All arms here use 5,118 examples and 320 updates. Tiny differences are descriptive; no seed-stability claim is made.}
\label{tab:mixture_matched}
\begin{tabular}{lrrrr}
\toprule
Model & Personal.$-$Bal. & Personal.$-$Prop. & Shared$-$Bal. & Shared$-$Prop. \\
\midrule
Qwen2.5-7B & $-0.255$ & $-0.464$ & $+0.340$ & $+0.130$ \\
Qwen3-4B & $-0.489$ & $+0.221$ & $-1.000$ & $-0.290$ \\
Qwen1.5-7B & $-0.897$ & $-0.720$ & $+0.164$ & $+0.341$ \\
Llama-3.1-8B & $-0.286$ & $+0.488$ & $-0.636$ & $+0.137$ \\
Llama-3.2-3B & $-0.238$ & $+0.196$ & $+0.055$ & $+0.489$ \\
Gemma-2-9B & $+0.550$ & $+0.328$ & $-0.038$ & $-0.260$ \\
\bottomrule
\end{tabular}
\end{table}

Personalized Top3 exceeds Balanced only on Gemma-2 (one of six models); Shared does so on three. Both exceed Proportional on four. Even the favorable Gemma comparison combines a $+2.312$ pp Math-family change with a $-0.659$ pp Code-family change relative to Balanced. Ranking individual categories therefore does not guarantee that their combination retains every benefit. Differences from the earlier 1,706-example single-category training configurations also change the training budget and cannot establish category synergy or interference.

For Qwen2.5, Shared exceeds Complete by $+11.429$ pp on MATH Medium and $+9.542$ pp on MATH High. Displayed scores are 66.7 versus 55.2 and 34.7 versus 25.2; Base scores are 64.8 and 31.3. These are improvements over both references, but the gaps to Base are much smaller. Shared is also lower than Complete on HumanEval ($-2.439$ pp) and MBPP-Simple ($-1.556$ pp), and its Code-family mean is lower by about 0.979 pp. This is evidence for selected-task gains, not uniform replacement of full-corpus training.

\clearpage
\subsection{Generation-budget sensitivity and resource interpretation}
\label{app:mixture_gsm_budget}
GSM8K is additionally evaluated at a 1,024-token cap for Base and every trained arm on the same 1,319 questions. Table~\ref{tab:mixture_gsm_base} shows that a larger budget especially benefits the Qwen2.5 and Qwen3 bases. Across the reported model--arm means, 34 of 36 absolute GSM8K scores increase, but trained-arm gains relative to the \emph{same-budget} Base fall from 16 of 30 positive contrasts to 6 of 30 (Table~\ref{tab:mixture_gsm_delta}). Ten positive contrasts become negative. These results remain separate from the primary eleven-task Overall; a single-task summary in the long-budget source plots is not that Overall.

\begin{table}[H]
\centering\small
\caption{\textbf{GSM8K Base sensitivity to generation budget.} Positive-arm counts cover the five trained recipes per backbone.}
\label{tab:mixture_gsm_base}
\begin{tabular}{lrrrr}
\toprule
Model & Base, 256 & Base, 1,024 & Change (pp) & Positive arms: 256$\to$1,024 \\
\midrule
Qwen2.5-7B & 76.346 & 91.054 & $+14.708$ & $5\to0$ \\
Qwen3-4B & 76.649 & 93.025 & $+16.376$ & $5\to1$ \\
Qwen1.5-7B & 59.591 & 63.154 & $+3.563$ & $1\to0$ \\
Llama-3.1-8B & 72.176 & 78.317 & $+6.141$ & $5\to5$ \\
Llama-3.2-3B & 72.328 & 76.801 & $+4.473$ & $0\to0$ \\
Gemma-2-9B & 84.685 & 85.444 & $+0.758$ & $0\to0$ \\
\bottomrule
\end{tabular}
\end{table}

\begin{table}[H]
\centering\small
\caption{\textbf{GSM8K changes from same-budget Base (pp).} Each cell gives the 256-token change followed by the 1,024-token change. These are separately generated evaluations, not forced continuations.}
\label{tab:mixture_gsm_delta}
\setlength{\tabcolsep}{3pt}
\begin{tabular}{lrrrrr}
\toprule
Model & Personal. & Shared & Balanced & Proport. & Complete \\
\midrule
Qwen2.5 & $7.51\to-1.36$ & $9.93\to-1.97$ & $7.88\to-1.21$ & $8.57\to-2.58$ & $7.51\to-2.12$ \\
Qwen3 & $10.46\to-0.45$ & $10.01\to0.30$ & $9.33\to-0.08$ & $9.33\to-0.38$ & $13.12\to-2.35$ \\
Qwen1.5 & $-2.12\to-5.61$ & $-2.81\to-6.52$ & $-2.81\to-6.97$ & $-2.43\to-5.53$ & $0.23\to-2.50$ \\
Llama-3.1 & $3.49\to3.18$ & $2.73\to0.99$ & $3.34\to2.43$ & $3.26\to2.27$ & $5.99\to3.64$ \\
Llama-3.2 & $-2.73\to-5.08$ & $-5.31\to-8.72$ & $-6.29\to-6.60$ & $-3.64\to-6.44$ & $-23.12\to-22.37$ \\
Gemma-2 & $-2.27\to-1.67$ & $-2.20\to-1.90$ & $-2.73\to-2.05$ & $-1.59\to-1.29$ & $-2.05\to-1.74$ \\
\bottomrule
\end{tabular}
\end{table}

For Qwen2.5 Shared, the GSM8K advantage over Complete is $+2.426$ pp at 256 tokens but only about $+0.15$ pp at 1,024. All five Qwen2.5 arms fall below the long-budget Base; all five Llama-3.1 arms remain above it. The source audit also reports shorter mean responses for every trained arm at both caps, so brevity alone does not determine gain direction. The two budgets use separate sampled generations at temperature 0.2; identical seeds do not guarantee identical prefixes. Actual stop reasons and per-item tokenized-prompt hashes were not archived. The result supports budget sensitivity, without attributing every recovered answer to truncation or identifying a causal mechanism.

The compact arms use 14.23\% of Complete's examples and updates. In the documented training audit, their final-training time ratios range from 13.77\% to 16.26\%, sequence-token ratios from 12.53\% to 15.73\%, and supervised-token ratios from 12.20\% to 15.64\%. These are recorded training-resource summaries, not three-seed uncertainty estimates or end-to-end savings. They exclude category screening, recipe search, and evaluation. Compact versus full comparisons also change token exposure and schedule length; the scores alone do not establish full-corpus overfitting or harmful categories.

\subsection{Further confirmation}
Independent validation should fix a selection objective and category rule using separate selection data, then test the frozen recipes against equal-budget balanced and proportional samples on held-out tasks or questions. Paired seed records and token- or update-matched controls would help distinguish repeatability from training-budget effects. An equivalence or non-inferiority claim additionally requires a prespecified margin, uncertainty on the paired contrast, and full discovery-cost accounting. These stronger claims are left to future work.

\clearpage
\subsection{Complete task scores}
\clearpage

\clearpage
\subsection{Task-level screening against constituent categories and Complete}
\label{app:mixture_screening}
Let $\bar S_{b,t}^{a}$ denote the three-seed mean percentage score of arm $a$ for backbone $b$ on task $t$.
For backbone $b$, task $t$, and a set of $k$ categories $C$ ($k=2$ or $3$), the broader screen requires
\begin{equation}
\bar S_{b,t}^{\mathrm{mix}(C)} > \max_{c\in C}\bar S_{b,t}^{c},
\qquad
\bar S_{b,t}^{\mathrm{mix}(C)} > \bar S_{b,t}^{\mathrm{Complete}}.
\label{eq:mixture_screen}
\end{equation}
Here $\mathrm{mix}(C)$ is one jointly trained model; each constituent is trained separately. Complete uses 35,974 examples. To additionally require every constituent to improve on Base, we impose
\begin{equation}
\bar S_{b,t}^{\mathrm{Base}} < \min_{c\in C}\bar S_{b,t}^{c}.
\label{eq:mixture_base_positive}
\end{equation}
The tables retain all broader-screen passes and bold mixture scores that also meet Equation~\ref{eq:mixture_base_positive}. Scores are reported three-seed means, as confirmed by the authors. Top2 uses seeds 20260730, 20260729, and 20260731; the earlier Top3 round uses 20260730, 20260729, and 20270731. Broader decisions use the reports' full-precision comparisons; tables show two decimals. Per-seed arrays were not supplied for independent reconstruction or uncertainty estimation.

For three categories, the broader screen retains $21/132$ comparisons: ten Personalized and eleven Shared results across $17$ model--task pairs; nineteen exceed Base. The additional condition retains $18/132$ across $14$ model--task pairs. Excluding GSM8K gives $20/120$ broader passes and $17/120$ stricter passes.

\begin{table}[H]
\centering\footnotesize
\caption{\textbf{All 21 broader three-category passes.} Best single is the highest constituent score. Scores are percentages; $\Delta$Base is in pp. Bold mixture scores mark the 18 passes for which every constituent also exceeds Base.}
\label{tab:three_category_screen}
\tabsetupplain
\setlength{\tabcolsep}{3pt}\renewcommand{\arraystretch}{1.05}
\begin{tabular}{@{}lllrrrr@{}}
\toprule
\headrow \hd{Backbone} & \hd{Task} & \hd{Recipe} & \hd{Best single} & \hd{Mixture} & \hd{Complete} & \hd{$\Delta$Base} \\
\midrule
Qwen2.5 & ARC-C & $\catlabel{\catIS}{Imp}\,+\,\catlabel{\catStr}{Str}\,+\,\catlabel{\catHash}{Hash}$ & 59.64 & \textbf{61.52} & 59.13 & $+10.07$ \\
 & ARC-C & $\catlabel{\catIS}{Imp}\,+\,\catlabel{\catHash}{Hash}\,+\,\catlabel{\catSort}{Sort}$ & 59.64 & \textbf{61.18} & 59.13 & $+9.73$ \\
 & ScienceQA & $\catlabel{\catIS}{Imp}\,+\,\catlabel{\catStr}{Str}\,+\,\catlabel{\catHash}{Hash}$ & 70.59 & \textbf{71.45} & 70.64 & $+3.42$ \\
 & ScienceQA & $\catlabel{\catIS}{Imp}\,+\,\catlabel{\catHash}{Hash}\,+\,\catlabel{\catSort}{Sort}$ & 70.59 & \textbf{71.18} & 70.64 & $+3.15$ \\
 & GSM8K$^{\dagger}$ & $\catlabel{\catIS}{Imp}\,+\,\catlabel{\catHash}{Hash}\,+\,\catlabel{\catSort}{Sort}$ & 84.08 & \textbf{86.28} & 83.85 & $+9.93$ \\
 & MATH-H & $\catlabel{\catIS}{Imp}\,+\,\catlabel{\catHash}{Hash}\,+\,\catlabel{\catSort}{Sort}$ & 30.15 & 34.73 & 25.19 & $+3.44$ \\
 & FinQA & $\catlabel{\catIS}{Imp}\,+\,\catlabel{\catStr}{Str}\,+\,\catlabel{\catHash}{Hash}$ & 19.33 & \textbf{19.95} & 16.59 & $+1.50$ \\
 & FinQA & $\catlabel{\catIS}{Imp}\,+\,\catlabel{\catHash}{Hash}\,+\,\catlabel{\catSort}{Sort}$ & 19.33 & \textbf{19.51} & 16.59 & $+1.06$ \\
 & MedCalc & $\catlabel{\catIS}{Imp}\,+\,\catlabel{\catStr}{Str}\,+\,\catlabel{\catHash}{Hash}$ & 11.00 & \textbf{12.82} & 11.91 & $+6.64$ \\
\addlinespace[2pt]
Qwen3 & LegalBench & $\catlabel{\catSeq}{Seq}\,+\,\catlabel{\catHash}{Hash}\,+\,\catlabel{\catMath}{Math}$ & 84.67 & \textbf{85.26} & 84.25 & $+1.95$ \\
 & LegalBench & $\catlabel{\catIS}{Imp}\,+\,\catlabel{\catHash}{Hash}\,+\,\catlabel{\catSort}{Sort}$ & 84.67 & \textbf{85.49} & 84.25 & $+2.19$ \\
\addlinespace[2pt]
Qwen1.5 & ARC-C & $\catlabel{\catIS}{Imp}\,+\,\catlabel{\catHash}{Hash}\,+\,\catlabel{\catSort}{Sort}$ & 57.17 & \textbf{57.34} & 54.69 & $+2.47$ \\
 & ScienceQA & $\catlabel{\catMath}{Math}\,+\,\catlabel{\catHash}{Hash}\,+\,\catlabel{\catGR}{Graph}$ & 69.69 & \textbf{69.74} & 66.37 & $+2.25$ \\
 & LegalBench & $\catlabel{\catMath}{Math}\,+\,\catlabel{\catHash}{Hash}\,+\,\catlabel{\catGR}{Graph}$ & 46.83 & 46.89 & 46.71 & $-0.30$ \\
 & MATH-M & $\catlabel{\catIS}{Imp}\,+\,\catlabel{\catHash}{Hash}\,+\,\catlabel{\catSort}{Sort}$ & 25.71 & \textbf{26.67} & 25.71 & $+5.71$ \\
 & MATH-H & $\catlabel{\catMath}{Math}\,+\,\catlabel{\catHash}{Hash}\,+\,\catlabel{\catGR}{Graph}$ & 8.40 & \textbf{8.78} & 7.63 & $+1.91$ \\
 & MBPP-Simple & $\catlabel{\catIS}{Imp}\,+\,\catlabel{\catHash}{Hash}\,+\,\catlabel{\catSort}{Sort}$ & 54.47 & \textbf{56.03} & 52.92 & $+4.67$ \\
\addlinespace[2pt]
Llama-3.2 & HumanEval & $\catlabel{\catIS}{Imp}\,+\,\catlabel{\catHash}{Hash}\,+\,\catlabel{\catSort}{Sort}$ & 51.83 & 54.27 & 47.56 & $0.00$ \\
\addlinespace[2pt]
Gemma-2 & ARC-C & $\catlabel{\catIS}{Imp}\,+\,\catlabel{\catStr}{Str}\,+\,\catlabel{\catMath}{Math}$ & 60.67 & \textbf{61.26} & 57.00 & $+4.52$ \\
 & FinQA & $\catlabel{\catIS}{Imp}\,+\,\catlabel{\catStr}{Str}\,+\,\catlabel{\catMath}{Math}$ & 21.27 & \textbf{21.62} & 18.36 & $+2.12$ \\
 & MedCalc & $\catlabel{\catIS}{Imp}\,+\,\catlabel{\catHash}{Hash}\,+\,\catlabel{\catSort}{Sort}$ & 15.00 & \textbf{15.27} & 15.00 & $+2.18$ \\
\bottomrule
\end{tabular}
\end{table}

\noindent\footnotesize
$\dagger$ GSM8K uses the 256-token primary protocol. The source lacks the full set of constituent single-category scores at 1,024 tokens, so this pass is not established under the larger budget. All other tasks retain their original evaluation settings.
\normalsize

The stricter condition excludes Qwen2.5 Shared on MATH-H, Qwen1.5 Personalized on LegalBench, and Llama-3.2 Shared on HumanEval. In the latter two cases, the mixture itself is below or tied with Base. On Qwen2.5 MATH-H, the mixture improves on Base even though its constituents individually decline, illustrating a pattern excluded by the additional condition.

\clearpage
\begin{table}[H]
\centering\footnotesize
\caption{\textbf{All 11 broader two-category passes.} Top2 retains the two leading categories in each backbone's earlier Overall ranking. Bold mixture scores mark the nine passes for which both constituents exceed Base. Other columns follow Table~\ref{tab:three_category_screen}.}
\label{tab:two_category_screen}
\tabsetupplain
\setlength{\tabcolsep}{3pt}\renewcommand{\arraystretch}{1.10}
\begin{tabular}{@{}lllrrrr@{}}
\toprule
\headrow \hd{Backbone} & \hd{Task} & \hd{Recipe} & \hd{Best single} & \hd{Mixture} & \hd{Complete} & \hd{$\Delta$Base} \\
\midrule
Qwen2.5 & ARC-C & $\catlabel{\catIS}{Imp}\,+\,\catlabel{\catStr}{Str}$ & 59.64 & \textbf{60.75} & 59.13 & $+9.30$ \\
 & ScienceQA & $\catlabel{\catIS}{Imp}\,+\,\catlabel{\catStr}{Str}$ & 70.59 & \textbf{70.77} & 70.64 & $+2.74$ \\
 & GSM8K$^{\dagger}$ & $\catlabel{\catIS}{Imp}\,+\,\catlabel{\catStr}{Str}$ & 84.00 & \textbf{84.84} & 83.85 & $+8.49$ \\
\addlinespace[2pt]
Qwen3 & LegalBench & $\catlabel{\catSeq}{Seq}\,+\,\catlabel{\catHash}{Hash}$ & 84.67 & \textbf{85.08} & 84.25 & $+1.78$ \\
 & MBPP-Simple & $\catlabel{\catSeq}{Seq}\,+\,\catlabel{\catHash}{Hash}$ & 78.60 & 79.38 & 77.82 & $-3.89$ \\
\addlinespace[2pt]
Qwen1.5 & ARC-C & $\catlabel{\catMath}{Math}\,+\,\catlabel{\catHash}{Hash}$ & 57.17 & \textbf{57.76} & 54.69 & $+2.90$ \\
 & ScienceQA & $\catlabel{\catMath}{Math}\,+\,\catlabel{\catHash}{Hash}$ & 69.69 & \textbf{69.96} & 66.37 & $+2.47$ \\
 & MATH-H & $\catlabel{\catMath}{Math}\,+\,\catlabel{\catHash}{Hash}$ & 7.25 & \textbf{8.40} & 7.63 & $+1.53$ \\
\addlinespace[2pt]
Llama-3.1 & ARC-C & $\catlabel{\catSort}{Sort}\,+\,\catlabel{\catIS}{Imp}$ & 52.05 & \textbf{52.30} & 49.57 & $+2.05$ \\
 & ScienceQA & $\catlabel{\catSort}{Sort}\,+\,\catlabel{\catIS}{Imp}$ & 68.17 & \textbf{68.21} & 65.83 & $+1.26$ \\
\addlinespace[2pt]
Llama-3.2 & HumanEval & $\catlabel{\catSort}{Sort}\,+\,\catlabel{\catDP}{DP}$ & 54.27 & 55.49 & 47.56 & $+1.22$ \\
\bottomrule
\end{tabular}
\end{table}

\noindent\footnotesize
$\dagger$ The same GSM8K budget qualification applies. The Qwen2.5 Top2 score is below the same-budget Base at 1,024 tokens; the screening conditions cannot be checked there without the constituent scores.
\normalsize

For two categories, the broader screen retains $11/66$ comparisons across five backbones, ten of which exceed Base. The additional condition retains $9/66$, excluding Qwen3 on MBPP-Simple and Llama-3.2 on HumanEval. Excluding GSM8K gives $10/60$ broader passes and $8/60$ stricter passes. These fractions have different candidate sets and constituent thresholds from the three-category screen; they do not compare interaction strengths across mixture sizes. Across both mixture sizes, the 27 stricter passes cover 16 distinct model--task pairs, five backbones, and nine evaluation configurations.

The additional condition is checked against the released one-decimal Base-relative task changes. A displayed 0.0 or $-0.0$ does not establish an exact tie. The only such value here is Llama-3.2's \catlabel{\catDP}{DP} result on HumanEval, whose \catlabel{\catSort}{Sort} constituent already declines; therefore no retained-count decision depends on an unresolved zero. All retained constituents have strictly positive displayed gains.

Two Qwen1.5 examples satisfy all conditions. On ARC-C, \catlabel{\catMath}{Math} (57.17\%) and \catlabel{\catHash}{Hash} (56.83\%) each exceed Base (54.86\%); their Top2 mixture reaches 57.76\%, above both constituents and Complete (54.69\%). On MBPP-Simple, \catlabel{\catIS}{Imp}, \catlabel{\catHash}{Hash}, and \catlabel{\catSort}{Sort} score 54.47\%, 54.09\%, and 54.47\%, each above Base (51.36\%); Shared reaches 56.03\%, versus Complete's 52.92\%. The margins over the best constituent are 0.60 and 1.56 pp. Both mixtures also exceed the existing Balanced and Proportional controls, although those 5,118-example controls are larger than Top2.

All selections are exploratory. Category rankings used the same test suite, and the task-level screen was applied after observing results. Single-category training uses 1,706 examples and 107 updates; Top2 uses 3,412 and 214; Top3 uses 5,118 and 320; Complete uses 35,974 and 2,249. Token exposure is also unmatched. Consequently, exceeding constituent models does not isolate category synergy from training amount, and exceeding Complete does not establish optimality or statistical significance. Independent recipe selection and matched-budget confirmation remain future work.

The accompanying CSVs preserve the broader passes, constituent Base-relative changes, and stricter subsets. A supplied Python script reproduces this audit and flags unresolved rounding boundaries without reconstructing per-seed results.
\clearpage
\section{Exploratory Audit of \catlabel{\catIS}{Direct Implementation} and QA Transfer}
\label{app:imp_qa_audit}
This appendix examines a possible interpretation of the \catlabel{\catIS}{Imp} response in RQ2: \catlabel{\catIS}{direct implementation and utilities (Imp)} may help a model apply stated conditions and variable relationships more faithfully. The analysis describes a secondary transfer phenomenon and its possible contributors. It does not identify a causal mechanism.

\subsection{Audit scope and comparative gains}
The audit covers one documented training seed, 20260730, with six backbones and twelve checkpoints per backbone: Base, ten single-category training configurations, and the balanced mixture. It uses 1,172 ARC-Challenge and 2,224 ScienceQA questions, yielding $72\times(1{,}172+2{,}224)=244{,}512$ saved choice predictions. Their selected options and correctness agree with the saved candidate scores and gold answers under the original scoring rule. These records supplement the reported three-seed summaries in the main analysis; they do not supply the missing seed-level variation. All averages below are descriptive means across the six fixed backbones.

Table~\ref{tab:imp_qa_contrasts} separates improvement over Base from advantage over other code categories. \catlabel{\catIS}{Imp} improves both benchmarks on every backbone in this audit. Its mean advantage over the other nine single-category training configurations is smaller and has exceptions. The other-nine comparator is the arithmetic mean of nine independently trained arms, not a voting ensemble; the balanced mixture is excluded from that mean.

\begin{table}[!htbp]
\centering\small
\caption{\textbf{Single-seed \catlabel{\catIS}{Imp} contrasts.} Accuracy differences in pp for seed 20260730. The final row averages the six backbones; it is not a training-replication estimate.}
\label{tab:imp_qa_contrasts}
\setlength{\tabcolsep}{6pt}
\begin{tabular}{@{}lrrrr@{}}
\toprule
& \multicolumn{2}{c}{ARC-Challenge} & \multicolumn{2}{c}{ScienceQA}\\
\cmidrule(lr){2-3}\cmidrule(l){4-5}
Backbone & \catlabel{\catIS}{Imp}$-$Base & \catlabel{\catIS}{Imp}$-$Other nine & \catlabel{\catIS}{Imp}$-$Base & \catlabel{\catIS}{Imp}$-$Other nine\\
\midrule
Qwen2.5-7B & $+8.19$ & $+2.28$ & $+2.56$ & $+0.33$\\
Qwen3-4B & $+6.23$ & $+0.84$ & $+5.53$ & $+1.51$\\
Llama-3.1-8B & $+1.79$ & $+0.36$ & $+1.21$ & $-0.04$\\
Gemma-2-9B & $+3.92$ & $+0.52$ & $+2.07$ & $+0.11$\\
Llama-3.2-3B & $+3.41$ & $+0.08$ & $+1.21$ & $-0.08$\\
Qwen1.5-7B & $+1.79$ & $-0.24$ & $+2.38$ & $+0.39$\\
\midrule
Mean & $+4.22$ & $+0.64$ & $+2.50$ & $+0.37$\\
\bottomrule
\end{tabular}
\end{table}

\subsection{Illustrative cases and the scoring protocol}
\catlabel{\catIS}{Direct-implementation} training includes tasks that turn a natural-language specification into variable bindings, threshold checks, and aggregation. For example, training item \texttt{Algorithm\_10021\_C} maps shipping type and quantity to a price, changes the rate at a ten-item threshold, and sums the costs. Such operations motivate the hypothesis that training can help apply existing knowledge to a question's conditions. Table~\ref{tab:imp_qa_cases} gives selected answer changes compatible with this interpretation, together with a factual improvement and a relational regression. Cases were selected after inspecting outcomes; their frequency cannot be inferred from this table.

ARC-Challenge and ScienceQA score each complete candidate answer, including its leading space, by mean token log probability and select the highest-scoring option. ScienceQA uses text and available hints without images. The template is fixed across arms within each backbone, and no explanation is generated. Consequently, these gains cannot arise directly from shorter generated explanations avoiding truncation or answer-extraction failures. This check does not exclude changes in option-length preference, wording preference, or likelihood calibration.

The reasoning descriptions in Table~\ref{tab:imp_qa_cases} are analyst interpretations of the question and selected answer, not decoded reasoning traces. Some changes occur near a decision boundary: the average-speed example has an \catlabel{\catIS}{Imp} top-two score gap of only 0.0116 mean log-probability units. Other inspected positives involving salinity and net force have gaps of 0.00375 and 0.0128. A tied-score case is also present in the audit, so unique correctness under the recorded tie-breaking rule alone is not strong evidence of acquired knowledge.
\FloatBarrier

\begin{table}[!htbp]
\centering\small
\caption{\textbf{Selected improvements and counterexamples.} Answers are paraphrased from saved predictions. R/F descriptions indicate the analyst's interpretation. These illustrative cases do not establish a mechanism or a prevalence estimate.}
\label{tab:imp_qa_cases}
\setlength{\tabcolsep}{5pt}\renewcommand{\arraystretch}{1.14}
\begin{tabularx}{\textwidth}{@{}p{0.25\textwidth}YY@{}}
\toprule
Backbone / item & Question and Base$\rightarrow$\catlabel{\catIS}{Imp} answer & Interpretation and comparison\\
\midrule
Llama-3.1 / ARC\newline\texttt{arc\_00372} & One hour at 80 km/h, then one hour at 100 km/h: $80\rightarrow90$ km/h. & Corrected equal-duration average speed; all other nine arms and the balanced mixture are wrong.\\
Qwen2.5 / ARC\newline\texttt{arc\_00564} & A substance expands and then melts when heated: initial state liquid$\rightarrow$solid. & Corrected initial-state inference; all other nine arms and the balanced mixture are wrong.\\
Llama-3.2 / ScienceQA\newline\texttt{scienceqa\_02187} & Given dominant $E$, recessive $e$, and genotype $ee$: brown$\rightarrow$red eyes. & Corrected rule application; all other nine arms and the balanced mixture are wrong.\\
Qwen2.5 / ARC\newline\texttt{arc\_00821} & \catlabel{\catIS}{Imp} correctly selects that plants existed before coal and oil formed; Base is wrong. & Factual improvement; all other nine single-category training configurations are wrong.\\
Llama-3.1 / ScienceQA\newline\texttt{scienceqa\_02144} & Both journeys take five hours: 75 miles$\rightarrow$60 miles as the faster journey. & Relational regression: Base is correct and \catlabel{\catIS}{Imp} is wrong.\\
Qwen2.5 / ScienceQA\newline\texttt{scienceqa\_01144} & Identify an object that is not a mineral: skull$\rightarrow$quartz. & Classification regression: Base and all other nine arms are correct.\\
\bottomrule
\end{tabularx}
\end{table}

\subsection{Do additional gains concentrate in relational questions?}
We inspect this prediction with a fixed sample of 80 questions per benchmark. Questions are sorted by the SHA256 hash of \texttt{imp-blind-20260924|task|question-id}, and the first 80 are retained, without selecting by \catlabel{\catIS}{Imp} correctness. A single AI annotator labels the question text as explicit relation/rule application (R), factual/definitional knowledge (F), language-specific rules (L), or other/ambiguous (O). Labels and reasons are retained, but there is no independent human adjudication. This is an exploratory, post hoc analysis, not a preregistered or independently verified double-blind study.

For question $i$, let $y_{bci}\in\{0,1\}$ be correctness for backbone $b$ and arm $c$. We first average each question's \catlabel{\catIS}{Imp} advantage over the six fixed backbones:
\[
d_i=\frac{100}{6}\sum_{b=1}^{6}\left(y_{b,\catlabel{\catIS}{Imp},i}-\frac{1}{9}\sum_{c\ne\catlabel{\catIS}{Imp}}y_{bci}\right),
\]
where the inner sum includes only the other nine single-category training configurations. We then compare $\overline d_R-\overline d_F$, resampling questions within each group for 20,000 bootstrap replicates (resampling seed 20260924). The percentile intervals quantify question-sampling uncertainty conditional on these models and labels. They do not include training randomness, annotation error, or uncertainty from choosing the hypothesis after observing the results.

\begin{table}[!htbp]
\centering\small
\caption{\textbf{Exploratory question-type comparison.} R/F gains are \catlabel{\catIS}{Imp} minus the other-nine mean, in pp. The intervals apply to R$-$F, not to training-seed variability.}
\label{tab:imp_qa_question_types}
\setlength{\tabcolsep}{6pt}
\begin{tabular}{@{}lrrrrl@{}}
\toprule
Benchmark & R/F $n$ & R gain & F gain & R$-$F & 95\% interval\\
\midrule
ARC-Challenge & 26/41 & $+0.28$ & $+2.89$ & $-2.61$ & $[-4.76,-0.54]$\\
ScienceQA & 20/15 & $+0.65$ & $+1.48$ & $-0.83$ & $[-3.95,+2.04]$\\
\bottomrule
\end{tabular}
\end{table}

This sample does not support the prediction that \catlabel{\catIS}{Imp}'s additional gains concentrate in relational questions: the ARC-Challenge difference is in the opposite direction, and the ScienceQA interval spans zero. The groups are small, the R definition excludes language-specific rules, and questions are not matched for difficulty or Base accuracy. The result therefore constrains the proposed explanation without establishing the absence of any benefit to rule application.

ScienceQA's full evaluated subset contains 1,043 natural-science, 1,056 language-science, and 125 social-science questions. Its total score should not be equated with natural-science reasoning. Grouping all natural-science questions by their original skill metadata provides a complementary check (Table~\ref{tab:imp_qa_skills}). Physical-relation and experimental-constraint questions show small advantages over the other categories while declining relative to Base. Thus, relative category preference can coexist with reduced absolute performance on a proposed mechanism-relevant subset.

\begin{table}[!htbp]
\centering\small
\caption{\textbf{ScienceQA natural-science skill groups.} Descriptive six-backbone mean changes in pp, using all questions in each indicated metadata group. ``Remaining'' questions are not a pure factual control.}
\label{tab:imp_qa_skills}
\setlength{\tabcolsep}{5pt}
\begin{tabularx}{\textwidth}{@{}Yrrr@{}}
\toprule
Skill group & $n$ & \catlabel{\catIS}{Imp}$-$Base & \catlabel{\catIS}{Imp}$-$Other nine\\
\midrule
Physical relations (e.g., speed, force, energy) & 141 & $-0.35$ & $+0.64$\\
Experimental constraints & 61 & $-1.37$ & $+0.82$\\
Genotype, phenotype, and dominance rules & 105 & $+2.38$ & $+0.67$\\
Inherited/acquired trait evidence & 113 & $+1.33$ & $+0.02$\\
Remaining natural-science questions & 623 & $+4.07$ & $+0.52$\\
\bottomrule
\end{tabularx}
\end{table}
\FloatBarrier

\subsection{Training content, response style, and open questions}
The 1,706 \catlabel{\catIS}{Imp} examples include aggregation (509), simple conversion (401), validation (368), mapping/filtering (167), library/API use (138), formatting (69), file-system operations (30), and date/time operations (24). These labels do not isolate a cognitive operation. Some formatting examples apply department and tenure thresholds, while some prompts already provide example implementations. Moreover, Python AST conditionals occur in 969/1,706 \catlabel{\catIS}{Imp} examples (56.8\%), compared with 1,605/1,706 \catlabel{\catDP}{dynamic-programming} examples (94.1\%) and 1,572/1,706 \catlabel{\catGR}{graph} examples (92.1\%). The records therefore do not support the simple explanation that \catlabel{\catIS}{Imp} contains a larger share of examples with conditionals; structural counts do not measure the quality of rule-application supervision.

The arms match 1,706 examples and 107 updates, but not token exposure. For Qwen2.5, \catlabel{\catIS}{Imp} uses 643,839 total input tokens and 208,684 supervised tokens, compared with 772,561/284,522 for \catlabel{\catDP}{dynamic programming} and 980,483/326,472 for \catlabel{\catGR}{graph} training. These are overlapping total and supervised counts, not quantities to add. Content, length, and supervision density remain coupled in the current comparison.

Generated-response records also show task-dependent style changes. For Qwen3, mean visible GSM8K length decreases from 181.5 to 163.7 tokens, while FinQA increases from 40.2 to 43.8 and MedCalc from 39.4 to 51.7. In one complete GSM8K response, \catlabel{\catIS}{Imp} applies a stated attendance fraction once, correcting Base's repeated application; in another, both responses derive the same four quantities but \catlabel{\catIS}{Imp} corrects their sum. These selected examples compare only Base with \catlabel{\catIS}{Imp} and cannot establish an \catlabel{\catIS}{Imp}-specific advantage over other categories. Visible lengths are reconstructed from saved text and do not recover original decoding stop reasons.

Taken together, the audit documents consistent Base-relative gains on these two QA benchmarks and examples compatible with improved application of stated conditions. It does not establish relational selectivity or a shared mechanism across backbones. Isolating that explanation would require training subsets annotated for the proposed operations, comparisons with matched token exposure and update counts, repeated training, and evaluation held out from hypothesis and data selection. We leave those controlled causal tests to future work.

The accompanying \texttt{data/imp\_qa\_audit/} directory preserves the summary tables, question sample and labels, training-budget and structure summaries, and saved case records used in this appendix. Its README distinguishes these derived records from the full prediction archive and from the main three-seed score summaries.
\clearpage
\section{Audit of QA Coverage and Backbone-Dependent Responses}
\label{app:finding1_audit}
This appendix examines the scope of RQ1 and possible contributors to its task-dependent responses. It uses saved outputs to distinguish task composition, changes in solutions, and evaluation-interface effects. The main results retain the reported three-seed aggregates; this complementary audit covers seed 20260730 only and does not independently verify the other two runs or estimate training variability.

\subsection{Records and protocol consistency}
The audited run contains six Base checkpoints, sixty single-category adapters, and six balanced-mixture adapters. All 66 training records document 1,706 examples, 107 optimizer steps, and the same LoRA settings. Evaluation contracts, templates, scoring populations, and decoding parameters match across the twelve arms within each backbone. Category token exposure is not matched. These checks support within-run comparisons; they do not isolate training-content effects from all other sources of variation.

QA counts and denominators agree across the saved metrics and 360 task-output files. The three mathematics configurations contain 121,392 saved outputs and 111,276 Base--trained question pairs; the three code configurations contain 57,528 outputs and 52,734 pairs. Paired outcomes are denoted CC (both correct), CW (Base correct, trained wrong), WC (Base wrong, trained correct), and WW (both wrong), according to the original scorer. The same Base response and question recur across category comparisons, so summed transition counts are checkpoint--question comparisons, not independent observations. Descriptive category means below exclude the balanced mixture, whose records are retained separately.

\subsection{How broadly do the QA gains extend?}
Table~\ref{tab:f1_qa_composition} recomputes task-family changes from full-precision correct-count records. All sixty single-category comparisons improve the five-task mean, but only 35 improve every constituent task individually. Removing ARC-Challenge and ScienceQA leaves 45 positive comparisons. Llama-3.1's remaining-three mean is close to zero, and Qwen1.5's is negative, whereas the other four backbones retain positive means in all ten categories.

\begin{table}[!htbp]
\centering\small
\caption{\textbf{QA composition sensitivity in seed 20260730.} Positive-arm counts and descriptive changes in pp. QA5 weights ARC-Challenge, FinQA, LegalBench, MedCalc, and ScienceQA equally; QA3 retains FinQA, LegalBench, and MedCalc. Changes average ten single-category training configurations per backbone.}
\label{tab:f1_qa_composition}
\setlength{\tabcolsep}{6pt}
\begin{tabular}{@{}lrrrr@{}}
\toprule
& \multicolumn{2}{c}{Positive arms} & \multicolumn{2}{c}{Mean change (pp)}\\
\cmidrule(lr){2-3}\cmidrule(l){4-5}
Backbone & QA5 & QA3 & QA5 & QA3\\
\midrule
Qwen2.5-7B & 10/10 & 10/10 & $+2.79$ & $+1.85$\\
Qwen3-4B & 10/10 & 10/10 & $+2.39$ & $+0.77$\\
Llama-3.1-8B & 10/10 & 5/10 & $+0.55$ & $+0.004$\\
Gemma-2-9B & 10/10 & 10/10 & $+2.26$ & $+1.95$\\
Llama-3.2-3B & 10/10 & 10/10 & $+1.45$ & $+0.86$\\
Qwen1.5-7B & 10/10 & 0/10 & $+0.47$ & $-0.57$\\
\midrule
All comparisons & 60/60 & 45/60 & $+1.65$ & $+0.81$\\
\bottomrule
\end{tabular}
\end{table}

Removing ARC-Challenge alone leaves 57/60 positive comparisons; removing ScienceQA alone leaves 56/60. Removing FinQA, LegalBench, or MedCalc individually leaves 60/60 in each case. Thus, the universal direction of the five-task mean partly depends on task composition, although substantial positive transfer remains beyond ARC-Challenge and ScienceQA. These counts are not rounding artifacts. They describe the specified benchmark configurations and do not establish uniform gains across QA domains or a particular reasoning mechanism. ScienceQA's evaluated subset also includes language- and social-science questions (Appendix~\ref{app:imp_qa_audit}).

\subsection{GSM8K: shorter responses accompany opposite score changes}
The single-seed GSM8K directions match RQ1: every category improves Qwen2.5, Qwen3, and Llama-3.1, while every category lowers Gemma-2, Llama-3.2, and Qwen1.5. Nevertheless, all sixty arms have shorter mean visible responses than their corresponding Base (Table~\ref{tab:f1_gsm_length}). Output shortening alone therefore does not distinguish the positive and negative responses.

\begin{table}[!htbp]
\centering\small
\caption{\textbf{GSM8K score and length changes.} Seed 20260730; $n=1{,}319$ questions per arm. Trained scores and lengths average ten category arms, not repeated training runs. Visible tokens are obtained by retokenizing saved text.}
\label{tab:f1_gsm_length}
\setlength{\tabcolsep}{6pt}
\begin{tabular}{@{}lrrrr@{}}
\toprule
Backbone & Mean $\Delta$ (pp) & Positive/negative & Base tokens & Trained tokens\\
\midrule
Qwen2.5-7B & $+7.09$ & 10/0 & 193.3 & 148.7\\
Qwen3-4B & $+3.71$ & 10/0 & 181.5 & 168.6\\
Llama-3.1-8B & $+2.84$ & 10/0 & 177.1 & 170.6\\
Gemma-2-9B & $-2.24$ & 0/10 & 100.0 & 89.0\\
Llama-3.2-3B & $-2.27$ & 0/10 & 175.6 & 160.9\\
Qwen1.5-7B & $-5.91$ & 0/10 & 169.4 & 113.7\\
\bottomrule
\end{tabular}
\end{table}

Replaying the archived GSM8K numeric scoring rule reproduces all 94,968 saved correctness flags. A separate sensitivity analysis takes the last explicit answer marker among \texttt{answer is}, \texttt{\#\#\#}, and boxed answers, falling back to the original rule if none exists. Applied uniformly to Base and every trained arm, it preserves all sixty single-category gain directions. However, it changes 440 originally correct records to incorrect and restores none. Some outputs literally end with the placeholder \texttt{The answer is N.}, for which the original rule can recover a numeric answer elsewhere in the response. The alternative is therefore a sensitivity check, not a validated correction or evidence that extraction errors are absent.

The case review identifies both kinds of scoring mismatch. Qwen2.5 Base correctly calculates $8.75-(-6.25)=15$ on \texttt{gsm8k\_00589}, but numbers in an unfinished closing summary interfere with extraction. Conversely, Qwen3 Base does not finish the requested calculation on \texttt{gsm8k\_00150}, yet a step number is extracted as the reference answer, 4. Original scores are retained for both cases. MATH-M and MATH-H use the original symbolic/numeric scoring outcomes; this audit does not independently implement a replacement symbolic scorer.

\subsection{Code failures include semantic and interface changes}
On HumanEval, Qwen3 has 154 CW and 36 WC comparisons across its ten categories, whereas Gemma-2 has 85 CW and 129 WC comparisons. Table~\ref{tab:f1_code_failures} shows that aggregate score changes can conceal different failure profiles. An assertion failure records a failed test, not a diagnosis that every such failure has the same semantic cause.

\begin{table}[!htbp]
\centering\small
\caption{\textbf{HumanEval transitions and recorded failure types.} Seed 20260730; sums over ten categories, each evaluated on 164 questions. The last four columns partition CW comparisons. Prompt-helper and indentation entries are compatibility diagnoses, not verified recoverable scores.}
\label{tab:f1_code_failures}
\setlength{\tabcolsep}{5pt}
\begin{tabular}{@{}lrrrrrr@{}}
\toprule
Backbone & CW & WC & Assertion & Helper & Indentation & Other\\
\midrule
Qwen3-4B & 154 & 36 & 131 & 10 & 0 & 13\\
Gemma-2-9B & 85 & 129 & 53 & 5 & 27 & 0\\
\bottomrule
\end{tabular}
\end{table}

For HumanEval/50, a helper supplied in the prompt is sometimes omitted when the evaluator extracts a complete target function. Tests then fail because the helper is unavailable. Other Gemma-2 outputs provide a function body whose outer indentation is not restored correctly by the wrapper. Static checks show that adding the missing outer indentation can make such code compilable, but the repaired programs were not executed in this audit. A syntax repair therefore cannot be counted as a recovered correct answer. All original test outcomes remain unchanged.

\newpage
Shortening also occurs in both directions within the same code benchmark: 150/154 Qwen3 HumanEval regressions (97.4\%) and 33/36 improvements (91.7\%) have shorter visible outputs than Base. Nor does every backbone shorten code responses: Gemma-2's mean HumanEval visible length rises from 67.5 to 87.1 tokens while its extracted-code character count falls, partly because extraction can include prompt material. These are different measurements, and neither alone establishes why correctness changes.

\subsection{Illustrative cases and interpretation limits}
The audit uses fixed samples of 24 GSM8K pairs (one per backbone and CC/CW/WC/WW stratum) and 48 HumanEval pairs (two per stratum). Model/category names are concealed and A/B order is permuted before a single AI review; labels are frozen before unblinding. This is not independent human adjudication or a representative estimate of error-type prevalence. Table~\ref{tab:f1_cases} selects examples from those records to illustrate different sources of change.

\begin{table}[!htbp]
\centering\small
\caption{\textbf{Examples of solution and interface changes.} Seed 20260730; analyst interpretations of saved outputs. Category names identify the inspected arm and do not imply that the change is unique to that category.}
\label{tab:f1_cases}
\setlength{\tabcolsep}{5pt}\renewcommand{\arraystretch}{1.12}
\begin{tabularx}{\textwidth}{@{}p{0.29\textwidth}Y@{}}
\toprule
Backbone / arm / item & Observed change and interpretation\\
\midrule
Qwen1.5 / \catlabel{\catStr}{String}\newline\texttt{gsm8k\_00972} & Base confuses 40 oranges kept with the number sold; the trained response correctly gives 120 sold. Both answers are complete, and the correct response is shorter (117$\rightarrow$105 visible tokens).\\
Gemma-2 / \catlabel{\catSort}{Sorting}\newline\texttt{gsm8k\_01070} & The trained response halves the quantity of discounted butter twice, changing the correct answer 18 to 15. The response becomes longer (87$\rightarrow$112 tokens).\\
Qwen3 / \catlabel{\catIS}{Imp}\newline HumanEval/26 & Base removes every value occurring more than once; the trained code instead retains one copy of each value. This changes required semantics. The same regression occurs in all ten categories and the mixture on this item.\\
Gemma-2 / \catlabel{\catSeq}{Sequence}\newline HumanEval/159 & A shorter program correctly updates remaining inventory by subtracting the available amount sold. This is a successful relation update rather than a length-induced failure.\\
Gemma-2 / \catlabel{\catSort}{Sorting}\newline HumanEval/50 & The decoding formula is retained, but the prompt-provided helper is absent from the assembled test program. The recorded regression is compatible with a wrapper failure.\\
\bottomrule
\end{tabularx}
\end{table}

These observations motivate a hypothesis that code training changes how existing solution strategies apply constraints and produce outputs, with consequences that depend on the backbone and evaluation interface. Individual cases exhibit actual relation or program-logic changes, while other outcomes depend on extraction or packaging. The audit does not identify their population-level causal contributions, and brevity alone cannot account for both improvements and regressions.

Original generation token IDs and stopping reasons are unavailable. Retokenized lengths near the generation cap are diagnostic proxies, not confirmed truncations. Subsets defined by both models producing short or explicit answers condition on post-training outputs and cannot establish an effect with truncation held constant. Controlled prompt or budget interventions, uniform validation of revised scoring, and cross-seed item-level checks remain future work.

The accompanying \texttt{data/finding1\_audit/} directory preserves the derived QA, mathematics, code, and protocol tables and the sampled case records. The full original audit archive additionally contains raw outputs, evaluator source, and execution logs. This manuscript bundle does not claim to include that complete archive or the two additional seed records.
\end{document}